\pdfoutput=1
\documentclass[11pt]{article}

\usepackage[]{emnlp2023}

\usepackage{times}
\usepackage{latexsym}
\usepackage{lipsum}
\usepackage{times}
\usepackage{latexsym}
\usepackage{hyperref}
\usepackage{adjustbox}
\usepackage{booktabs}
\usepackage{multirow}

\usepackage{microtype}
\usepackage{arydshln}
\usepackage[noend]{algpseudocode}
\usepackage{color,soul}
\usepackage{xspace}
\usepackage{mathtools}
\usepackage{amssymb}
\usepackage{amsmath}
\usepackage{cleveref}
\crefname{section}{§}{§§}
\Crefname{section}{§}{§§}
\usepackage{footnote}
\usepackage{tablefootnote}
\usepackage{graphicx}
\usepackage{color, colortbl}
\usepackage{xstring}
\usepackage{comment}
\usepackage{adjustbox}
\usepackage{float}
\restylefloat{table}
\usepackage{graphicx}
\usepackage{array,booktabs,makecell}
\usepackage{multirow}
\usepackage{graphicx}
\usepackage{appendix}
\usepackage{enumerate}
\usepackage{arabtex}

\usepackage{array}
\newcolumntype{H}{>{\setbox0=\hbox\bgroup}c<{\egroup}@{}}
\usepackage{xcolor}
\usepackage{color, colortbl}
\usepackage{utf8}
\usepackage{cancel}
\usepackage{ulem,xpatch}
\usepackage[T1]{fontenc}
\usepackage[utf8]{inputenc}
\usepackage{microtype}
\newcommand\blfootnote[1]{%
  \begingroup
  \renewcommand\thefootnote{}\footnote{#1}%
  \addtocounter{footnote}{-1}%
  \endgroup
}
\usepackage[T1]{fontenc}

\usepackage[utf8]{inputenc}
\usepackage{ascii}
\usepackage{newunicodechar}
\newunicodechar{☺}{\SOH}
\newunicodechar{☻}{\STX}
\newunicodechar{♥}{\ETX}
\newunicodechar{♣}{\ENQ}
\newunicodechar{♠}{\ACK}
\newunicodechar{•}{\BEL}
\newunicodechar{○}{\HT}
\usepackage{microtype}

\def \ourmodel{JASMINE} 
\title{\ourmodel: Arabic GPT Models for Few-Shot Learning}

\author{\normalsize El Moatez Billah Nagoudi$^{\lambda,\star}$~Muhammad Abdul-Mageed$^{\lambda,\xi,\star}$~  AbdelRahim Elmadany$^{\lambda}$  \\
\normalsize\textbf{Alcides Alcoba Inciarte}$^{\lambda}$~\textbf{Md Tawkat Islam Khondaker}$^{\lambda}$ \\
\normalsize $^{\lambda}$ Deep Learning \& Natural Language Processing Group,
  The University of British Columbia\\\normalsize  $^{\xi}$Department of Natural Language Processing \& Department of Machine Learning, MBZUAI\\ %
  \texttt{\normalsize \{moatez.nagoudi,muhammad.mageed,a.elmadany\}@ubc.ca}}
  
\begin{document}
\setcode{utf8}
\maketitle
\pagenumbering{Arabic}

% \begin{abstract}
\section*{~~~~~~~~~~~~~~~~~~~~~~~~~~~~~Abstract}
Scholarship on generative pretraining (GPT) remains acutely Anglocentric, leaving serious gaps in our understanding of the whole class of autoregressive models. For example, we have little knowledge about the potential of these models and their societal impacts in diverse linguistic and cultural settings. We alleviate this issue for Arabic, a wide collection of languages and dialectal varieties with  $\sim450$ million population, by introducing~\ourmodel. ~\ourmodel~is a suite of powerful Arabic autoregressive Transformer language models ranging in size between $300$ million-$6.7$ billion parameters pretrained on a large and diverse dataset ($\sim235$GB of text). We also carefully design and release a comprehensive benchmark for both automated and human evaluation of Arabic autoregressive models, with coverage of potential social biases, harms, and toxicity. Using our novel benchmark, we evaluate~\ourmodel~extensively showing powerful performance intrinsically as well as in few-shot learning on a wide range of NLP tasks.
We aim to responsibly release our models and evaluation benchmark with interested researchers, along with code for experimenting with them.
~\blfootnote{ $^{\star}${Authors contributed equally.}} 

% \end{abstract}
% Auto-regressive, generative models have recently resulted in significant progress in English-based NLP. It is not clear, however, how such models could fare in other languages such as these with complex morphology. We introduce {\bf A}rabic {\bf P}re{\bf t}raining, a class of Arabic-based pre-training with three main variants of ...

% \input{abstract}\label{sec:abs}

%------------------------------------------

%%%%%%%%%%%%%%%%%%%%%%%%%%%%%%%%%%%%%%%%%%%%%%%%%%%
\section{Introduction}\label{intro}
Recent work in generative pretraining~\cite{radford2019language,brown2020language,lieber2021jurassic,chowdhery2022palm,zhang2022opt,smith2022using,scao2022bloom,thoppilan2022lamda,hoffmann2022training} has shown that autoregressive models perform well on language tasks using in-context learning, without finetuning or gradient updates. %That is, these decoder-only models are good one- and few-shot learners and may perform above chance in the zero-shot scenario. 
This in-context learning approach allows models to perform new tasks with only simple instructions and a few optional examples, which can be further improved by model adaptation through prompt tuning~\cite{lester2021power}. %This progress caused a shift in focus from the pretrain-then-finetune scenario typical in fully supervised learning to that of few-shot learning. The effectiveness of in-context learning was also further improved by model adaptation through prompt tuning~\cite{lester2021power}. %Introduction of the GPT-2~\cite{radford2019language} and GPT-3 models~\cite{brown2020language} triggered significant interest in autoregressive models in the community, with several other models following. These include, for example, GPT-J-6B~\cite{gpt-j}, Jurassic~\cite{lieber2021jurassic}, Gopher~\cite{rae2021scaling}, PaLM~\cite{chowdhery2022palm}, Megatron-Turing Natural Language Generation (MT-NLG-530)~\cite{smith2022using}, LaMDA~\cite{thoppilan2022lamda}, Chinchilla~\cite{hoffmann2022training},  OPT~\cite{zhang2022opt}, and BLOOM~\cite{scao2022bloom}.
In spite of this progress, autoregressive pretrained Transformer language models of significant size remain largely \textit{anglocentric}. This makes it difficult to bring more diverse voices to the table. Nor is it clear if multilingual models such as BLOOM~\cite{scao2022bloom}, where model capacity is split across a large number of languages and language-specific data are neither sufficiently large nor diverse, can allow equitable understanding of these models in languages other than English. %In addition, these multilingual models are usually pretrained with less, and less diverse, language-specific data than what may be possible through dedicated (i.e., language-specific) models. Under these current conditions, 
It is also not possible to study the capabilities of these models in particular linguistic environments (e.g., languages of rich morphology, of diglossic nature, and/or with a large number of dialects such as Arabic) and diverse cultural backgrounds (e.g., African, Asian, Latin American). This situation also deprives non-English communities of the rich array of benefits language model technology can bring as its full potential and emerging capabilities~\cite{wei2022emergent} are unlocked. Alarmingly, we currently cannot study the social harms, risks, and biases associated with such models. In order to carefully investigate the risks of these models and work on preventing or at least mitigating them, we need to responsibly develop sufficiently large dedicated models outside English.% without development of sufficiently large dedicated models outside English. Studying social biases and ethical considerations may also be challenging using multilingual models such as Bloom, possibly due to the cross-lingual interaction and potential knowledge transfer within the model.

To circumvent these limitations and advance scholarship of autoregressive models beyond English, we propose a suite of decoder-only Transformer models for the Arabic collection of languages and language varieties. Our suite of models, dubbed~\ourmodel, come in four different architectures that range in size from $300$ million to $6.7$ billion parameters. Motivated by recent findings as to the impact of pretraining data size \textit{vis-à-vis} model size~\cite{hoffmann2022training, falcon2023}, we carefully curate a large dataset ($\sim235$GB of text) of high-quality text to pretrain~\ourmodel. Our dataset is also diverse (e.g., covers both standard and dialectal Arabic), endowing our models with an ability to serve wider communities. %Because Arabic is a collection of languages rather than a single monolithic variety, as explained (also see~\cref{sub_sec:arabic}), we believe our diverse data lends our models the potential to serve wider communities. 

Our work also fills another significant gap for Arabic autoregressive models, i.e., that of an evaluation benchmark. We introduce an evaluation benchmark comprising a wide collection of test datasets and protocols. Using our benchmark, we evaluate~\ourmodel~extensively both \textit{intrinsically}  (using perplexity) and \textit{extrinsically} (e.g., on few-shot settings). Our evaluation demonstrates the superiority of~\ourmodel~compared to available baselines. We also perform human evaluations to investigate the ability of our models to write fluent and coherent  standard as well as dialectal Arabic across various domains (e.g., news, literary, Twitter). Our evaluations reveal that our~\ourmodel~models posses powerful representations, allowing them to excel in few-shot learning and produce outputs that can be identified by humans only at chance level. %. We also find that humans annotators perform at chance level when trying to tease apart natural and~\ourmodel-generated texts, which further demonstrates the superiority of ~\ourmodel.
Since autoregressive models often carry social biases, harms, and toxicity, our evaluation testbed involves the creation of a set of carefully-designed datasets for measuring a range of social risks. Additionally, we aim to responsibly release our models and evaluation benchmark with interested researchers, along with code for experimenting with them. 

To summarize, we offer the following contributions: \textbf{(1)} We develop~\ourmodel, a suite of four autoregressive language models for Arabic, ranging in size between $300$ million to $6.7$ billion parameters pretrained with a diverse dataset. % of $412$ GB of text. %Our pretraining data is diverse in terms of domain and coverage of Arabic varieties. 
\textbf{(2)} We evaluate~\ourmodel~extensively, introducing a comprehensive evaluation benchmark for a wide range of NLP tasks. We demonstrate~\ourmodel's ability to write fluent language and learn well in-context across rich contexts in few-shot settings. \textbf{(3)} Our evaluation benchmark involves the creation and release of datasets for investigating potential social biases, harms, and toxicity. Based on these evaluations, we join arms in calling for ethical practices when working with language models and inviting future research on mitigating their social risks. \textbf{(4)}~We aim to responsibly and gradually release our models with interested researchers, along with code for experimenting with them, hoping our work will trigger applications and further research in understanding autoregressive models outside English.

The rest of the paper is organized as follows:  We introduce \ourmodel~in Section~\ref{sec:argpt3_model}, describe our evaluation strategies in Section~\ref{sec:Eval_strategies}, and our evaluation benchmark in Section~\ref{sec:Eval}. In Section~\ref{sec:human_eval}, we offer human evaluations of model output. Section~\ref{sec:soc-bias-analysis} is an analysis of social bias in the model, and Section~\ref{sec:lit} is about related work. We conclude in Section~\ref{sec:conclusion}. %We now introduce our \ourmodel~models.

\section{\ourmodel}\label{sec:argpt3_model}

\subsection{Arabic}\label{sub_sec:arabic}
\textit{Arabic} is a collection of languages and language varieties, some of which (e.g., Moroccan Arabic and Egyptian Arabic) are not mutually intelligible. \textit{Classical Arabic (CA)} is the variety used in old Arabic poetry and the Qur'an, and is employed side by side with other varieties to date. \textit{Modern Standard Arabic (MSA)} is a more modern variety~\cite{badawi1973levels} of Arabic that is usually used in pan-Arab media, government, and formal education across the Arab world. \textit{Dialectal Arabic (DA)} is the term used to refer to Arabic dialects. Dialects are sometimes defined regionally (e.g., Gulf, Levantine, Nile Basin, and North African~\cite{Habash:2010:introduction,mageed2015Dissert}), but also at the country or even province levels (e.g., ~\cite{Bouamor:2018:madar,mageed2020microdialect,mageed:2020:nadi,abdul-mageed-etal-2021-nadi, abdul2022nadi}). We now introduce~\ourmodel.

\subsection{(Pretraining) Data}\label{sub_sec:lm_data}

%We describe the dataset and procedure we employ to pretrain~\ourmodel.  As mentioned, Arabic is a collection of languages and language varieties, involving a classical variety (CA), a modern variety (MSA), and several dialects (DA). 

Our dataset is linguistically diverse, covering all categories of Arabic (i.e., CA, DA, and MSA), as we will now describe.

\noindent\textbf{CA Data.} 
We use the  Open Islamicate Texts Initiative (OpenITI) corpus  (v1.6)~\cite{nigst2020openiti}.\footnote{
We exclude a random sample of $1$K books from  OpenITI for later use in evaluating~\ourmodel~ perplexity (see \cref{subsec:ppl}).} OpenITI  contains $11,195$ premodern Islamic books mainly collected from Shamela Liberay,\footnote{\href{https://shamela.ws}{https://shamela.ws}.} Al-Jami Al-Kabir collection (JK),\footnote{\href{http://kitab-project.org/docs/openITI}{http://kitab-project.org/docs/openITI}.} books digitized by Jordanian publisher Markaz Al-Turāth, and the Shia  Library.\footnote{\href{https://shiaonlinelibrary.com}{https://shiaonlinelibrary.com}.} 
\textbf{MSA Data.}  We use $\sim$$223$ GB of MSA text ($23.7$ billion tokens) from the following sources: AraNews\textsubscript{v2}~\cite{nagoudi2020machine}, %\footnote{We upgrade \newcite{nagoudi2020machine}'s data by crawling news Articles from a new set of $150$ newspapers from across the whole Arab world.} 
El-Khair~\cite{elkhair-2016}, Gigaword,\footnote{\href{https://catalog.ldc.upenn.edu/LDC2009T30}{https://catalog.ldc.upenn.edu/LDC2009T30}.} OSCAR~\cite{suarez2019asynchronous}, OSIAN~\cite{zeroual2019osian},  Wikipedia Arabic, and Hindawi Books.\footnote{\href{https://www.hindawi.org/books/}{https://www.hindawi.org/books}.} We also extract the Arabic part of the multilingual Colossal Clean Crawled Corpus (mC4)~\cite{xue2020mt5} and clean it (see ~\cref{subsec:pre_data} for cleaning procedure). We call the extracted portion AraC4 (more details are in Appendix~\ref{app:arac4}). \textbf{Dialectal Data (DA).} 
We use a corpus of $1.5$ billion Arabic tweets ($178$GB) randomly sampled from a large in-house dataset of $\sim13$~billion Arabic tweets. This dataset is used only for finetuning one of our models (see Section~\ref{sec:human_eval}), rather than pretraining.

\noindent\textbf{Data Distribution.}  We analyze the distribution of MSA vs. DA in both our AraC4 and Twitter collections using a SoTA binary classifier~\cite{abdul-mageed-etal-2021-arbert} (MSA vs. dialect, $\sim88\%$ F\textsubscript{1}) on a random sample of $100$ million samples from each. We find that our Twitter data involves $28.39$\% predicted dialect tweets and our AraC4 data involves $5.7\%$ predicted dialect sentences. We then run another SoTA country-level classifier~\cite{abdul-mageed-etal-2021-arbert} ($\sim40\%$ F\textsubscript{1}) on the predicted dialect portions from each dataset, finding that our Twitter data is more diverse than AraC4. For example, our classifier tags $80$\% of the predicted AraC4 dialects as Egyptian, 2.86\% as Bahraini, 1.85\% as Libyan, leaving other dialects to be only marginally represented. Refer to Table~\ref{tab:aragpt3_data} for more information about our pretraining data (e.g., size, number of tokens) and Table~\ref{tab:dia_distrib} for country-level predicted dialects from each of the datasets.

\subsection{Preprocessing and Vocabulary}\label{subsec:pre_data}
We clean our pretraining data  by removing HTML tags,  elongation, and hash signs. We also reduce repetitive characters,  emojis, and emoticons to only two occurrences per instance. Further, we replace URLs and user mentions with the \texttt{<URL>} and \texttt{<USER>} strings. To create our vocabulary, we use a BPE-based tokenizer similar to GPT-2 \cite{radford2019language}, with a vocabulary of $64,000$ BPE tokens. Refer to Appendix~\ref{app:vocab} for more details.

% \usepackage{graphicx}
\begin{table}[t]
\centering
\resizebox{0.7\columnwidth}{!}{%
\begin{tabular}{lcHr}
\toprule
\textbf{Source}                          & \textbf{Size}  & \textbf{Sentences} & \textbf{Tokens} \\ \toprule

\small AraC4     &  \small $173$GB                   &  \small $xx$M           & \small $19.8$B                               \\
\small AraNews\textsubscript{v2} & \small $18.3$GB                   &\small  $4.26$M           & \small $1.8$B                               \\
\small El-Khair & \small $16$GB      & \small $42$M           & \small $1.6$B                              \\
\small  Hindawi\textsubscript{v2}     & \small $1.1$GB                   & \small $2.3$M            & \small $78.6$M                                 \\

\small  Gigawords    & \small $10$GB                    & \small $67.4$M           &\small  $1.1$B                              \\

\small  OSIAN     &\small  $2.8$GB                   &\small  $12$M           & \small $292.6$M                                \\

% OSCAR-MSA        & $31$GB                    & $67.3$M           & $3.4$B          \\
\small  OSCAR-Egy      & \small $32$MB & \small $101.9$M              &\small  $3.8$M                                  \\
\small  Wiki     &\small  $1.6$GB                   &\small  $12.5$M           & \small $156.5$M                                \\
 \hline
\small  \textbf{MSA-Total}                           & \small $222.8$GB & \textbf{$ \bf  cc$M}                      &\small  $23.7$B   

\\ \hline
\small  \textbf{CA} &\small  $12$GB                   & $xx$M           & \small $1.1$B                               \\\hline
\small  \textbf{MSA+CA} &\small  $243.8$GB                   & $xx$M           & \small $24.8$B                               \\\hline
\small  \textbf{Twitter } &\small  $178$GB                   & \small $1.5$B           & \small $21.9$B                               \\ \toprule
% \small  \textbf{ALL} & \small $ \bf 412.8$GB                   & $xx$M           & \small $\bf46.7$B
% \toprule

\end{tabular}%
}
\caption{Datasets used in~\ourmodel~models.}
\label{tab:aragpt3_data}
\end{table}
% ###################################

\subsection{Model Design and Implementation}\label{sub_sec:modeles}
%-------------------------------------
We exploit our diverse dataset to train four different variants of~\ourmodel, as follows:   \textbf{\ourmodel\textsubscript{350M}}, \textbf{\ourmodel\textsubscript{1.3B}},  \textbf{\ourmodel\textsubscript{2.7B}}, and \textbf{\ourmodel\textsubscript{6.7B}}.\footnote{The number of parameters is suffixed to model names.} We pretrain~\ourmodel~models for $500$k steps each using the autoregressive next-step prediction objective~\cite{radford2019language} and  the Transformer-based GPT-Neo~\cite{black2021gpt} replication of the GPT-3~\cite{brown2020language} architecture. %\hl{For all models, we use a sequence length of $2,048$ tokens and a batch size of $512$.  Our largest model has $13$ billion parameters, with  $40$ layers, an embedding dimension of $5,120$, and $40$ attention heads.} 
Details of the various architectures of~\ourmodel~are in Table~\ref{tab:jasmine_config}. 

% \usepackage{graphicx}
\begin{table}[!t]
\centering
 \renewcommand{\arraystretch}{1.2}
\resizebox{1\columnwidth}{!}{%
\begin{tabular}{lccccc}
\toprule
\textbf{Model}                          &  \textbf{Layers} & \textbf{Heads} & \textbf{Embed}& \textbf{Seq}  & \textbf{\# Parameters}\\  \midrule

\textbf{\ourmodel\textsubscript{350M} }&       $12$       &      $12$ & $768$          &     $2,048$    &    $350$M    \\ 
\textbf{\ourmodel\textsubscript{1.3B} }   &      $24$       &      $16$ & $2,048$          &     $2,048$    &    $1.3$B     \\
\textbf{\ourmodel\textsubscript{2.7B}}  &      $32$       &      $32$ & $2,560$          &     $2,048$    &      $2.7$B    \\
\textbf{\ourmodel\textsubscript{6.7B}  }   &      $32$       &      $32$ & $4,096$          &     $2,048$    &    $6.7$B    \\ 
%\textbf{\ourmodel\textsubscript{13B}  }   &       $40$       &      $40$ & $5,120$          &     $2,048$    &    $13$B    \\  
\toprule

% \textbf{AraGPT\textsubscript{10B} }   &                     &          &          &     &                     \\ \hline
% \textbf{Avg. PPL}                           &     &                   &    &   &    \\ \hline

\end{tabular}%
}
\caption{Parameter values for our~\ourmodel~models.}
\label{tab:jasmine_config}
\end{table}

% ###################################
% \textbf{AraGPT2\textsubscript{135M} }   &          $382.62$           & $463.63$ &   $71.30$      &                         \\
% \textbf{AraGPT2\textsubscript{370M} }       &    $217.00$      &                    $276.50$    &        $81.10$    &              \\
% \textbf{AraGPT2\textsubscript{792M}}  &         $639.91$        &         $499.11$       &      $71.72$   &        \\
% \textbf{AraGPT2\textsubscript{1.4B} }   &           $533.05$     &          $347.00$      &   $57.55$    &          \\  \cdashline{1-6}
% %%%%%%%%%%%%%%%%%%%%%%%%%%%%%%%%%%%%
% \textbf{mGPT\textsubscript{1.4B} }            &      $394.48$  &          $122.78$        &     $\underline{19.98}$   &         \\ \hline
% %%%%%%%%%%%%%%%%%%%%%%%%%%%%%%%%%%%%%

% \noindent\textbf{AraGPT\textsubscript{350M}}.

% \noindent\textbf{AraGPT\textsubscript{1.3B}}.

% \noindent\textbf{AraGPT\textsubscript{2.7B}}.
%%%%%%%%%%%%%%%%%%%%%%%%%%%%%%%%%%%%%%%%%%%%%%%%%%%
%%%%%%%%%%%%%%%%%%%%%%%%%%%%%%%%%%%%%%%%%%%%%%%%%%%

\section{Evaluation Strategies}\label{sec:Eval_strategies}

%Evaluating autoregressive (i.e., causal) language models  can be challenging~\cite{howcroft2020twenty}. Most work focuses either on intrinsic measures such as the perplexity or gauging extrinsic abilities of models on downstream tasks in few-shot settings without any previous finetuning (i.e., in-context learning)~\cite{brown2020language, zhang2022opt}. 
We follow previous literature~\cite{brown2020language,howcroft2020twenty,zhang2022opt} in evaluating our models extensively, under both intrinsic and extrinsic conditions as we now explain. % We explain our evaluations next. %three settings: (1) \textit{language modeling }(i.e., perplexity), (2) \textit{zero, one,  and few-shot}, and (3)\textit{ fine-tuning }settings.  

\noindent\textbf{Intrinsic Evaluation.}
\textit{Perplexity} (PPL) is a widely used metric that estimates how well a language model predicts a given text. %It is defined as the exponentiated average negative log-likelihood of a text (or sequence of tokens). That is, 
For a tokenized text $T = (w_{1}, w_{1}, ..., w_{n}) $, perplexity of $T$ is:
\begin{equation}
    \mathit{PPL (T)} = exp\{-\frac{1}{n}\sum_{i}^{n} \; log\ p_{0}(w_{i}| w_{<i})\}
\end{equation}
Where $log\ p_{0}(w_{i}| w_{<i})$ is the log-likelihood of the $i^{th}$ word conditioned on the previous words $w_{<i}$.

\noindent\textbf{Extrinsic Evaluation.}
We employ three settings: \noindent\textit{(1)~few-shot}, where a model is given $k$ examples describing the task at inference time as conditioning, but without updating the models' weights. \noindent\textit{(2)~one-shot}, which is the same as few-shot except that only one example is provided to the model (i.e., $k$=$1$). \noindent\textit{ (3)~zero-shot},  where no demonstrations are provided to the model (i.e., $k$=$0$).%, and the model is only given a natural language instruction describing the task.

% \subsection{Finetuning}
% We also evaluate our models under a full finetuning setting. In fact, we update the weights of our pretrained models by finetuning them on a set of natural language understanding 
% % and generation datasets.

\section{Evaluation Benchmark}\label{sec:Eval}

We evaluate \ourmodel~on $23$ different datasets, representing five different tasks: \textit{language modeling}, \textit{autocompletion}, \textit{commonsense inference,}  \textit{word manipulation}, and \textit{natural language understanding}. We now introduce each of these tasks along with related datasets.

% \textcolor{red}{Can we add MT?} 

\subsection{Language Modeling}\label{subsec:ppl}

 As explained, we calculate the perplexity of our models as intrinsic evaluation. Since there is no standard dataset for evaluating perplexity on Arabic texts, we create and release a new multi-domain dataset totaling $6$K documents extracted from six publicly available sources. These datasets are not in our pretraining and cover three Arabic varieties: MSA, dialect, and CA. We introduce each of them. \textbf{(1)~Arabic Wikipedia.} We select $1$K articles from  Arabic Wikipedia  (\textit{AraWiki}), published after October $2022$ to avoid leakage with our data.%\footnote{The Dec. $2021$ Wikipedia dump is in our pretraining.} 
 \textbf{(2) WikiLingua.} Introduced by~\newcite{ladhak-wiki-2020}, this resource contains article and summary pairs in $18$ languages, including Arabic, extracted from WikiHow.\footnote{\href{https://www.wikihow.com/}{https://www.wikihow.com/}.} 
We extract $1$K Arabic articles from the test set of WikiLingua.\footnote{\href{https://huggingface.co/datasets/GEM/wiki\_lingua}{https://huggingface.co/datasets/GEM/wiki\_lingua}.} \textbf{(3)~News Articles.} We collect $1$K news articles from $\sim100$ Arabic online sources.
The articles are not in our pretraining and cover different domains (e.g., culture, economy, politics, sports). \textbf{(4)~Watan2004.}  We select $1$K articles from an old dataset, Watan2004 (WT04)~\cite{abbas2011evaluation}. 
For dialectal and classical Arabic, we also extract a random $1$K articles from each of the following sources: \textbf{(5)~EgyWiki.}~Egyptian Arabic articles from Wikipedia dumps, and  \textbf{(6)~CA-Book.} Open Islamicate Texts Initiative (OpenITI) corpus~\cite{nigst2020openiti}. 
%WikiHow is a collaborative, high-quality resource of how-to guides on a diverse set of topics written by human authors. 
 %that are not in our pretraining data. %To avoid data leakage with our pretraining dataset, we only collect articles published after March $2022$. The Arabic news 
 %More details about our three perplexity datasets are in Appendix Z.

\noindent\textbf{Results.} Table~\ref{tab:aragpt3_ppl} shows the zero-shot BPE-token level perplexity of our~\ourmodel~models on the six datasets. We compare to the four AraGPT2 models proposed by~\newcite{antoun2021aragpt2} and mGPT~\cite{shliazhko2022mgpt} as baselines. Our~\ourmodel~models clearly outperform all baselines by a significant margin, with~\ourmodel\textsubscript{6.7B} reaching %$33.06$, $31.93$, and $16.81$ on the AraWiki, WikiLing, and AraNews test datasets, respectively. 
an average PPL of $42.25$. %We also note that although our largest model (i.e., \ourmodel\textsubscript{13B}) is still pretraining, it achieves a promising performance.%In addition, while our $6.7$B and $13$B models are still pretraining at early stages (i.e., $291$K and $57$K steps, respectively), they both achieve promising performance. For example, ~\ourmodel\textsubscript{6.7B} ($291$K steps) already outperforms all our baselines. 

\begin{table}[t]
\centering
 \renewcommand{\arraystretch}{1.2}
\resizebox{1\columnwidth}{!}{%
\begin{tabular}{lcccccc|c}
\toprule
\textbf{Model}    & \multicolumn{1}{c}{\textbf{AraWiki}} & \multicolumn{1}{c}{\textbf{WikiLing}} & \multicolumn{1}{c}{\textbf{AraNews}} & \textbf{WT04} & \textbf{EgyWiki} & \textbf{Op-ITI}     & \multicolumn{1}{c}{\textbf{Avg.}} \\\toprule
\textbf{AraGPT2\textsubscript{135M} }     &$ 87.55$&$ 65.27 $&$ 34.22$&$  44.26      $&$ 368.71      $&$ 181.83 $&$ 119.50   $\\
\textbf{AraGPT2\textsubscript{370M} }     &$ 68.93$&$ 57.57 $&$ 27.53$&$ 38.26       $&$ 265.17      $&$ 133.25  $&$ 91.07    $\\
\textbf{AraGPT2\textsubscript{792M}}      &$ 51.37$&$ 49.43 $&$ 30.65$&$ 32.15       $&$ 395.67      $&$ 122.13  $&$ 103.08   $\\
\textbf{AraGPT2\textsubscript{1.4B} }     &$ 34.72$&$ 44.88 $&$ 27.59$&$ 26.90$&$ 289.91      $&$ 121.35  $&$ 82.85    $\\
\cdashline{2-8}
\textbf{mGPT\textsubscript{1.4B} } &   $ 394.48       $&$ 122.78$&$ 19.98$&$ 156.01 $&$ 141.78      $&$ 148.67  $&$ 164.37   $\\ \hline
\textbf{\ourmodel\textsubscript{350M} }  &$ 52.10 $&$ 49.02 $&$ 23.88$&$ 40.82       $&$ 182.45      $&$ 108.55  $&$ 72.02    $\\
\textbf{\ourmodel\textsubscript{1.3B} }  &$ 35.75$&$ 36.08 $&$ 18.45$&$ 27.65       $&$ 106.33      $&$ 84.14   $&$ 48.78    $\\
\textbf{\ourmodel\textsubscript{2.7B}}    &$ 33.06$&$ 31.93 $&$ 16.81$&$ 24.73       $&$ 91.71       $&$ 81.98   $&$ 44.53    $\\
\textbf{\ourmodel\textsubscript{6.7B} }  
 & $\bf30.27$ & $ \bf31.21 $ & $ \bf16.12$ & $ \bf23.45$ & $ \bf87.35$ & $ \bf77.32$ & $\bf42.25$ \\

\toprule

\end{tabular}%
}
\caption{Results in the perplexity of our~\ourmodel~models on our language modeling benchmark. We compare to AraGPT2~\cite{antoun2020arabert} and mGPT~\cite{shliazhko2022mgpt}.} %$^\star$~\textbf{Our $13$B models is still pretraining}. We report \ourmodel\textsubscript{13B} at $220$K steps.}
\label{tab:aragpt3_ppl}
\end{table}

\subsection{Autocompletion}
The goal of autocompletion is to predict the last word for a given text. For this, we create a dataset totaling $15$K samples. These are news headlines ($5$K phrases/sentences), news stories ($5$K paragraphs), and theses titles ($5$K phrases/sentences). All samples are collected from diverse online sources. For example, the thesis titles cover domains such us~\small \<الإدارة>  \normalsize(management), \small \<علم النفس> \normalsize(psychology), and \small \<القانون> \normalsize(law).  For evaluation, we give~\ourmodel~a prompt (title or paragraph) without the last word and ask it to predict the masked word. We experiment with our models under zero-, one-, and few-shot settings. \textbf{Results.} Table~\ref{tab:complition_res_rev} shows results on the news title datasets, and we provide results for the two other autocompletion datasets in Table~\ref{tab:complition_res_rev_app}. From Table~\ref{tab:complition_res_rev} we can see that~\ourmodel~models perform best in all settings.\footnote{For this and upcoming experiments, we restrict evaluation to our smaller models (all or any of our $1.3$B-$6.7$B models) due to constraints on our computing resources.} We also observe that more demonstrations tend to help improve performance. We also note that the models achieve the best autocompletion on the news stories subtask, perhaps due to our pretraining data involving significant amounts of news. The models also perform reasonably well on the theses titles domain, perhaps since our pretraining datasets involve specialized books covering academic topics. We notice a drop in model performance under the $24$-shot setting, perhaps since few-shot learning can be sensitive to the order of the shots~\newcite{wei2021finetuned, brown2020language, lu-etal-2022-fantastically}. %\hl{We have to say why the 24-shots is lower than 16-(8) shots?}

%We also find that model performance does not necessarily increase as we increase the number of shots. The common wisdom for this behavior is that improvement with few-shot learning is model- and task-dependent, and is often sensitive to the order of the shots~\newcite{flan_wei, gpt_3, lu-etal-2022-fantastically}. We now discuss performance on each understanding task. 

% \input{Tables/complitaion_task_results_reverse.tex}

% \usepackage{multirow}

\begin{table}[t]
 \renewcommand{\arraystretch}{1.1}
 \resizebox{1\columnwidth}{!}{
\begin{tabular}{llcccccH}

\toprule
                       & \textbf{\small  Models }      & \textbf{0-shot} & \textbf{1-shot}   &\textbf{8-shots}     &\textbf{16-shots} & \textbf{24-shots} & \textbf{32-shot}      \\ \toprule

% ---------------------

\multicolumn{1}{c}{\multirow{8}{*}{\rotatebox[origin=c]{90}{\small \textbf{News Title}}}}

&\small\textbf{AraGPT2\textsubscript{135M} }&$ 11.13  $&$ 10.38    $&$ 12.47    $&$ 12.19 $&$ 12.82 $&$ 12.17    $\\
& \small\textbf{AraGPT2\textsubscript{370M} }&  $ 10.86  $&$ 11.42    $&$ 12.78    $&$ 13.77 $&$ 13.18 $&$ 14.43    $\\
&\small \textbf{AraGPT2\textsubscript{792M}} &  $ 13.61  $&$ 15.24    $&$ 16.74    $&$ 19.33 $&$ 14.44 $&$ 15.96    $\\
& \small\textbf{AraGPT2\textsubscript{1.4B} }&  $ 14.92  $&$ 15.22    $&$ 11.51    $&$ 17.00 $&$ 10.89 $&$ 16.45    $\\ \cdashline{3-8}
& \small\textbf{mGPT\textsubscript{1.3B} }& $ 12.80   $&$ 13.63     $&$ 10.32     $&$ 10.48  $&$ 10.34  $&$ 11.12 $ \\ \cline{2-7}
& \small\textbf{JASMINE\textsubscript{350M} } &$ 12.79  $&$ 13.39    $&$ 16.09    $&$ 18.04 $&\textbf{$16.67$}&$ 18.38    $\\
&\small \textbf{JASMINE\textsubscript{1.3B} } &$ 15.25  $&$ 16.13    $&$ 17.49    $&$ 20.98 $&$ 16.01 $&$ 19.36    $\\
& \small\textbf{JASMINE\textsubscript{2.7B}}  & \textbf{$15.88$} &\textbf{ $16.93$}   & \textbf{$17.57$}& \textbf{$23.13$}  &$ 15.82 $&\textbf{$\bf20.58$}    \\
& \small\textbf{JASMINE\textsubscript{6.7B}   }&    $\bf15.91$&$\bf17.44$&$\bf18.41$&$\bf24.10$&$\bf17.96$&$9.55$ \\          
% &\small\textbf{\ourmodel\textsubscript{13B}  }&$ -$&$- $&$- $&$ - $&$ - $&$ -$ \\ 
% $&$ \textbf{JASMINE\textsubscript{10B} }   $&$   $&$ $&$ $&$  $&$  $&$ $\\ \hline

% --------------------
\toprule

\end{tabular} }
\caption{Zero-, one-, and few-shot performance in F\textsubscript{1} on the news title  completion tasks.}\label{tab:complition_res_rev}
\end{table}

\subsection{Commonsense Inference}
%%%%%%%%%%%%%%%%%%%%%%%%%%%%%%%%%%%%%%
Since there is no Arabic  \textit{commonsense inference} evaluation dataset, we follow methods introduced by~\citet{swag} to create a new, high-quality Arabic commonsense collection using a random sample of $16,707$ examples from Arabic WikiHow. Each example has a context and a correct answer.\footnote{\href{https://www.wikihow.com/}{https://www.wikihow.com}} For each context, we create three generated answers using an adversarial approach. We refer to our new dataset as \textbf{AraSWAG} (\textbf{Ara}bic \textbf{S}ituations \textbf{W}ith \textbf{A}dversarial \textbf{G}enerations). We next provide a full explanation of it.

%in Appendix~\ref{app:araSwag} and Figure~\ref{fig:arswag_algorithm} in the same appendix.

\noindent\textbf{Initial Dataset Creation.} We randomly sample $10$K examples from Arabic WikiHow.\footnote{\href{https://www.wikihow.com/}{https://www.wikihow.com}} We then finetune AraT5~\citep{nagoudi-etal-2022-arat5} on the sampled examples separately, where we feed the model with the contexts in order to generate the endings. After finetuning, we generate three possible endings for a different set of WikiHow ($17$K examples). We generate the ending by setting top\textsubscript{k} = $50$ and top\textsubscript{p} = $0.95$ to mimic human-like writings. Therefore, our initial datasets contain one context and four endings (one \textit{real} and three \textit{generated}).

\noindent\textbf{Adversarial Dataset Creation.} To make the commonsense inference task more challenging, we follow~\cite {swag,hellaswag} and apply the adversarial filtering (AF) method on the initial dataset. Specifically, on each iteration, the dataset is randomly partitioned into $\mathcal{D}_{train}$ and $\mathcal{D}_{test}$ with a split of $8$:$2$. We then finetune a MARBERT~\citep{abdul-mageed-etal-2021-arbert} model in order to  classify endings as \textit{real} or \textit{generated} on $\mathcal{D}_{train}$. We evaluate the finetuned model on $\mathcal{D}_{test}$, then apply AF to replace easy-to-classify generations in $\mathcal{D}_{test}$ with newly generated endings using the finetuned AraT5. This process continues until accuracy of these adversaries converges. We observe that during convergence, the accuracy of MARBERT drops to $\sim30$\%. Finally, we randomly split the resulting \textbf{AraSWAG} dataset into training (Train=$14,288$), validation (Dev= $7,44$), and testing (Test=$1,675$) sets.

 %
%%%%%%%%%%%%%%%%%%%%%%%%%%%%%%%%%%%%%%%%%%%%%%%%%%%%%
% \subsubsection{Datasets Collection}
%\tawkat{Please describe how Arabic WikiHow and ActivityNet Captions dataset are prepared.}
%%%%%%%%%%%%%%%%%%%%%%%%%%%%%%%%%%%%%%%%%%%%%%%%%%%%%%
% \subsubsection{Initial Datasets Creation}
%%%%%%%%%%%%%%%%%%%%%%%%%%%%%%%%%%%%%%%%%%%%%%%%%%%%%%
 %We observe that performance significantly deteriorates on the adversarial dataset, compared to the initial dataset. This proves that AF creates a final dataset that is more challenging to the model and hence qualifies as the basis of a more interesting commonsense inference task. 
%%%%%%%%%%%%%%%%%%%%%%%%%%%%%%%%%%%%%%%%%%%%%%%%%%%%%%
% \input{Tables/Generated_examples_potery.tex}
% \input{Tables/new_potery_examples.tex}
% \input{Tables/new_potery_examples3.tex}
%%%%%%%%%%%%%%%%%%%% ArSWAG Figure %%%%%%%%%%%%%%%%%%%%%%%%%%%
\begin{figure}[]
    \centering
    \includegraphics[width=\columnwidth]{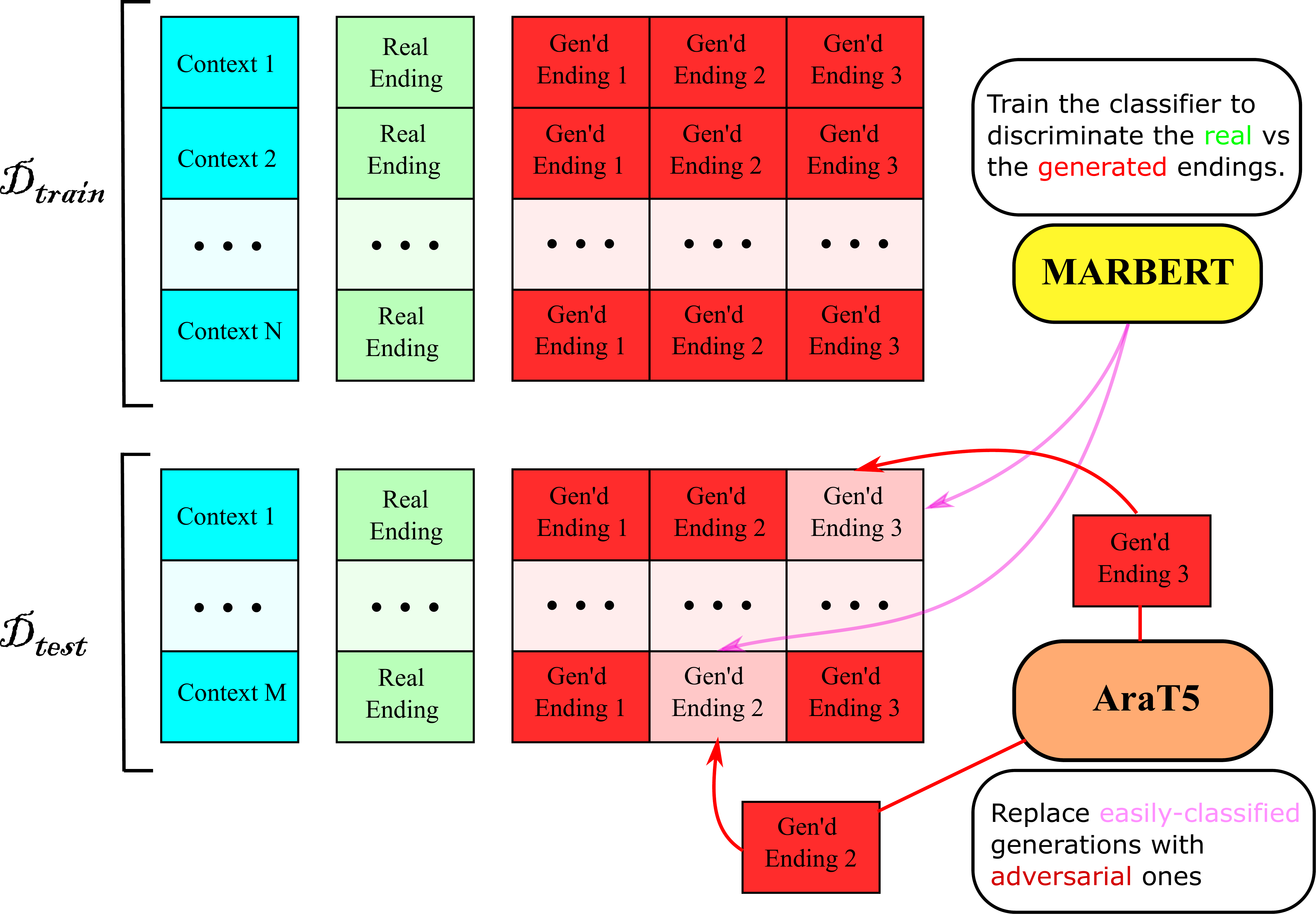}
    \caption{Overview of \textbf{AraSWAG} dataset creation. On
    each iteration, a new MARBERT is trained on a dummy
    training set $\mathcal{D}_{train}$ to identify \textcolor{pink}{\textit{easily-classified}} generated endings on the dummy test set $\mathcal{D}_{test}$. The finetuned AraT5 is used to replace \textcolor{pink}{\textit{easily-classified}} generated endings with \textcolor{red}{\textit{adversarial}} ones. This process is repeated iteratively to obtain a challenging dataset.} 
    \label{fig:arswag_algorithm}   
\end{figure}
%%%%%%%%%%%%%%%%%%%% ArSWAG Figure %%%%%%%%%%%%%%%%%%%%%%%%%%%
%\noindent\textbf{Evaluating~\ourmodel~on AraSWAG.} 
We use AraSWAG  to seed our $350$B,  $1.3$B, and $2.7$B~\ourmodel~models and the baselines with a context and four endings, one original (true) and three generated  (false) as explained. We then compute for each ending a \textit{language modeling score} (LMS), following~\citet{nadeem2021stereoset},\footnote{Refer to Appendix~\ref{app:araSwag} for details about LMS.} to identify whether it is \textit{related} to the seed context or not. We evaluate the likelihood of each candidate's ending conditioned on the context and choose the candidate with the highest \textit{LMS}. Table~\ref{tab:arswag_prompt} shows an example of a context and four endings from AraSWAG. \textbf{Results.} As Table~\ref{tab:arswag_result} shows, although our dataset is challenging,~\ourmodel\textsubscript{2.7B} significantly outperforms baselines ($37.18$ F\textsubscript{1}).% both in terms of accuracy and F\textsubscript{1}. 

\begin{table}[t]
\centering
 \resizebox{0.85\columnwidth}{!}{%
\begin{tabular}{l}
\toprule
\textbf{~~~~~~~~~~~~~~AraSwag Prompting Example} \\
\toprule 

\begin{tabular}[c]{@{}l@{}}\textbf{\colorbox{blue!15}{\small Context:}}

\small ~~~~\< احرصي على نظافتكِ الشخصية. احصلي على  >\\

\small ~~~~~~~~~~~~~~\<قسط كافِ من النوم. تناولي طعامًا صحيًا وابتعدي >\\ 

\small  ~~~~~~~~~~~~~~~~~~~~~~\< عن الوجبات السريعة. اشربي المزيد من المياه. >\\

\small ~~~~~~~~~~~~~~~~~~~~~~~\<مارسي التمارين الرياضية للبقاء بصحة جيدة. >\end{tabular} \\
\textbf{\colorbox{red!15}{\small Ending 1}:} \small\< تحلي باليقظة والحفاظ على نظام غذائي صحي >                                                                                                                                                                          \\
\textbf{\colorbox{green!15}{\small Ending 2}:}~~~\small\< وفي الأخير اشربي أكوابا من الماء  و السوائل >                                                                                                                                                \\
\textbf{\colorbox{red!15}{\small Ending 3}:} ~~~~~~~~~~~~\small\<  اعرفي ما يمكنكِ فعله لإنقاص وزنكِ>                                                                                                                                                  \\
\textbf{\colorbox{red!15}{\small Ending 4}:} ~~~~~~~~\small\< تحدثي إلى طبيبكِ بخصوص تغيير شكلكِ>     \\ \midrule

% \begin{tabular}[c]{@{}l@{}}\textbf{\colorbox{green!15}{Context:} \small\< و في الأخير اشربي أكوابا من الماء  و الكثير من السوائل >   } \end{tabular} \\ 
\rowcolor{green!15}
\textbf{Answer: } \begin{tabular}[c]{@{}l@{}}\small \< وفي الأخير اشربي أكوابا من الماء  و  السوائل >    \end{tabular}\\
\toprule
% \textbf{Text:} \small\< افهم الخلايا. ضع مفهومًا لنموذجك. فكر في المواد التي ستستخدمها >                                                                                                                                                      \\
% \textbf{{[}1:} \small\< رتب القوالب >                                                                                                                                                                                                      \\
% \textbf{{[}2:}\small \< اختبر كل خلاياك >                                                                                                                                                                                                  \\
% \textbf{{[}3:}\small \< نسق بياناتك >                                                                                                                                                                                                      \\
% \textbf{{[}4:} \small\< كن مبتكرًا >                                                                                                                                                                                                     \\
% \textbf{\colorbox{green!20}{Answer:}}                                                                                                                                                                                                                 \\ \toprule
\end{tabular}%
}
\caption{A context and four endings from  AraSWAG, with the second ending as a correct answer.} \label{tab:arswag_prompt} %after computing the LMS for each ending candidate.
\end{table}

\begin{table}[t]
\centering
\small
 \renewcommand{\arraystretch}{1.2}
\resizebox{0.63\columnwidth}{!}{%
\begin{tabular}{lcc}
\toprule

 \textbf{Models } &  \textbf{Acc}   &  F\textsubscript{1}   \\ \midrule

% & \textbf{AraGPT2\textsubscript{792M}} &  $ 22.93  $&$ 24.36    $&$ 24.36    $&$ 24.36 $&$ 24.24 $&$ 26.03    $\\
% & \textbf{AraGPT2\textsubscript{1.4B} }&  $ 23.58  $&$ 24.00    $&$ 24.60    $&$ 24.36 $&$ 23.52 $&$ 24.24    $\\ \cdashline{3-8}
% & \textbf{mGPT\textsubscript{1.3B} }& $ 24.42   $&$ \bf 26.51     $&$ \bf 26.51    $&$ 26.62  $&$ 24.24  $&$ 24.36 $ \\ \cline{2-8}
 \textbf{AraGPT2\textsubscript{135M} } &$   23.64$&$23.61 $ \\
  \textbf{AraGPT2\textsubscript{370M} } &$28.23$&$28.23 $ \\
 \textbf{AraGPT2\textsubscript{792M} } &$  32.59 $&$32.03 $ \\ 
 \textbf{AraGPT2\textsubscript{1.4B} } &$ 26.74  $&$26.75$ \\ \midrule
 \textbf{JASMINE\textsubscript{350M} } &$ 28.23 $&$28.23$\\
 \textbf{JASMINE\textsubscript{1.3B} } &$\ 35.28 $ &$ 35.26 $  \\
  \textbf{JASMINE\textsubscript{2.7B} } &$\bf 37.23 $ &  $\bf 37.18 $  \\
   % \textbf{JASMINE\textsubscript{6.7B} } &$\bf 35.28 $ &$ \bf 35.26 $  \\
  % \textbf{JASMINE\textsubscript{2.7B} } &$\bf - $ &$ - $  \\
% & \textbf{JASMINE\textsubscript{2.7B}}  & $ 24.36  $ &$ 24.36 $   & $ 24.36 $&$ 24.36 $&$ 24.36 $&$ 24.36 $    \\
% & \textbf{JASMINE\textsubscript{6.7B}  }&$ 24.60  $&$ 24.54 $&$ 24.54 $&$ \bf 27.46 $&$ \bf 26.15 $&$ \bf 26.99 $\\\hline
% % $&$ \textbf{JASMINE\textsubscript{10B} }   $&$   $&$ $&$ $&$  $&$  $&$ $\\ \hline
\toprule   
\end{tabular}

}
\caption{Performance on the \textbf{AraSWAG} dataset.}

\label{tab:arswag_result}
\end{table}

\subsection{Word Manipulation} \label{subsec:wc_scram}
%%%%%%%%%%%%%%%%%%%%%%%%%%%%%%%%
 
% \usepackage{multirow}
\begin{table}[t]
\renewcommand{\arraystretch}{1.2}
 \resizebox{0.98\columnwidth}{!}{%
\begin{tabular}{clHHHcccccc}
\toprule
\multicolumn{1}{l}{}      & \multicolumn{1}{c}{\textbf{Setting}} & \textbf{AraGPT2\textsubscript{135M} }       & \textbf{AraGPT2\textsubscript{370M} }       & \textbf{AraGPT2\textsubscript{792M}} & \textbf{AraGPT2\textsubscript{1.4B} }       & \textbf{mGPT\textsubscript{1.4B} }& \textbf{\ourmodel\textsubscript{350M} }& \textbf{\ourmodel\textsubscript{1.3B} }& \textbf{\ourmodel\textsubscript{2.7B}} & \textbf{\ourmodel\textsubscript{6.7B}}  \\
\toprule

\multicolumn{1}{c}{\multirow{5}{*}{QALB}}    & 0-shot    &&&& 0.10&0.11&0.10&0.10&0.11&\textbf{0.12}\\
\multicolumn{1}{c}{}      & 1-shot     & &&&1.11&1.63&1.14&1.67&\textbf{2.58}&1.94\\
\multicolumn{1}{c}{}      & 8-shots      & &&&0.92&1.41&2.5&3.70&5.49&\textbf{5.88}\\
\multicolumn{1}{c}{}      & 16-shots     & &&&1.72&2.72&4.27&\textbf{4.75}&4.24&4.37\\
\multicolumn{1}{c}{}       & 24-shots     & &&&1.19&1.35&2.51&3.87&4.25&\textbf{4.58}\\ \midrule \midrule

\multicolumn{1}{c}{\multirow{5}{*}{A1}}    & 0-shot    & 0.10 & 0.00  & 0.00   & 0.10  & 0.15 & 0.40   & 0.45  & \textbf{1.01} & 0.99  \\
\multicolumn{1}{c}{}      & 1-shot     & 0.12 & 0.23  & 0.80   & 1.77  & 0.30 & 0.96   &  2.28 & 2.03  & \textbf{2.57}   \\
\multicolumn{1}{c}{}      & 8-shots      & 0.00 & 0.00  & 0.55   & 0.00  & 0.60 & 1.56   & 2.88  & \textbf{4.48} &  4.28 \\
\multicolumn{1}{c}{}      & 16-shots     & 0.00 & 0.00  & 0.65   & 0.93  & 0.70 & 0.99   & 2.80  & 3.60 & \textbf{3.90}  \\
\multicolumn{1}{c}{}       & 24-shots     & 0.00 & 0.00  & 1.72   & 1.39  & 1.52 & 4.35   & 5.16  & 5.41 &  \textbf{5.63} \\ \midrule
 % & 32-shots     & 0.00 & 0.00  & 4.00   & 4.90  & \textbf{1.85}& 3.57   & 5.17  & \textbf{5.17} &   \\
\multicolumn{1}{c}{\multirow{5}{*}{A2}}    & 0-shot    & 0.80 & 0.87  & 0.10   & 0.25  & 0.97 & 1.63   & 1.27  & \textbf{2.68} & 1.89   \\
\multicolumn{1}{c}{}      & 1-shot     & 0.80& 0.87 & 1.83 & 3.91& 0.97 & 3.05 &  7.77  & 7.32  & \textbf{8.24}  \\
\multicolumn{1}{c}{}      & 8-shots      & 0.00 & 0.00  & 2.99   & 2.40  & 0.56 & 5.10   &8.32& \textbf{10.53}&  9.12  \\
\multicolumn{1}{c}{}      & 16-shots     & 0.00 & 0.00  & 2.91   & 1.80  & 0.00 & 5.88   & 7.55  & 8.49 & \textbf{10.57 } \\
\multicolumn{1}{c}{}      & 24-shots     & 1.64 & 0.00  & 4.00   & 1.30  & 1.49 & 7.04   & 9.72  & 10.70&   \textbf{11.94}\\  \midrule
  % & 32-shots     & 0.00 & 0.00  & 3.51   & 5.36  & 0.00 & 7.14   & 8.93  & \textbf{10.71}&   \\ \hline
\multicolumn{1}{c}{}      & 0-shot    & 1.78 & 0.60  & 0.81   & 0.76 & 1.89 & 5.21   & 5.99  &  7.30 & \textbf{7.93}    \\
\multicolumn{1}{c}{\multirow{5}{*}{RI}} & 1-shot     & 0.65& 0.92  & 5.73   & 7.18  & 0.00 & 7.78   & \textbf{11.34}& 9.48  & 9.95 \\
\multicolumn{1}{c}{}      & 8-shots      & 0.00 & 0.00  & 7.26   & 8.02  & 0.56 & 15.94  & \textbf{22.97}& 17.83 &  21.04 \\
\multicolumn{1}{c}{}      & 16-shots     & 2.90 & 0.00  & 3.14   & 2.44  & 2.08 & \textbf{14.77} & 12.90 & 11.96 & 11.36  \\
\multicolumn{1}{c}{}      & 24-shots     & 0.00 & 0.00  & 8.47   & 1.75  & 1.43& 8.33   & {16.93}& 10.94 & \textbf{17.82}  \\  \midrule
 % & 32-shots     & 2.78 & \textbf{0.00} & \textbf{12.45} & 13.33 & 3.70 & \textbf{17.78} & 16.33 & 12.00 &   \\ \hline
\multicolumn{1}{c}{\multirow{5}{*}{CL}}    & 0-shot    & 0.00 & 0.15  & 0.00   & 0.00  & \textbf{0.35}& 0.15   & 0.10  & 0.30  & \textbf{0.35}   \\
\multicolumn{1}{c}{}      & 1-shot     & 0.12 & 0.00  & 0.23   & 1.12  & 0.57 & 0.34   & 1.24  & \textbf{1.88} &   1.44 \\
\multicolumn{1}{c}{}      & 8-shots      & 0.00 & 0.00  & 1.58   & 5.18 & 1.63 & 3.00   & 4.37  & 3.34  &  \textbf{6.63} \\
\multicolumn{1}{c}{}      & 16-shots     & 0.00 & 0.00  & 4.21   & \textbf{7.62} & 1.94 & 3.95   & 4.59  & 3.60  & 6.54    \\
\multicolumn{1}{c}{}     & 24-shots     & 0.00 & 0.00  & 0.00   & 1.35  & 1.33 & 4.20   & 5.34  & \textbf{6.34} &  5.97 \\ \bottomrule  
 % & 32-shots     & 0.00 & 0.00  & 0.00   & 1.67  & 0.00 & 1.72   & 1.64  & \textbf{3.39} &      \\ \toprule      
\end{tabular}} \caption{Performance on the different word scrambling  tasks (F\textsubscript{1}). We exclude results for \textit{reversed words} from the table since, similar to GPT-3, the models did not predict any correct answers  (i.e., F\textsubscript{1}=0).}\label{tab:wc_scram}
\end{table}

% We follow the same five-word scrambling technique used in GPT-3~\newcite{radford2019language} to test the ability of our AraGPT models to learn how to fix the word manipulation from a few examples. The 
We test our~\ourmodel~models' ability to learn how to correct word-level errors (i.e., recover the original word) from a few examples. For this, we exploit one existing and one new dataset: \textbf{(i) Natural Spelling Errors}. We use QALB~\cite{zaghouani2014large}, a large manually-corrected collection of Arabic sentences. QALB covers a variety of types of errors, from which we extract $22.8$k words with spelling errors and errors in proper names. %proper names, word choice, morphology, and syntax. As in this task, we focus only on correcting errors at the word level (without), we extract $xx$k Arabic words from spelling and proper name errors categories. 
\begin{table}[t]
\centering
\small
 \renewcommand{\arraystretch}{1}
\resizebox{0.77\columnwidth}{!}{%
\begin{tabular}{crr}
\toprule

 \textbf{Manipulation } &  \textbf{\colorbox{green!20}{{Original   }}}   &  \textbf{\colorbox{red!20}{{Manipulated   }}}    \textbf{}  \\ \toprule
 \textbf{CL} &   \colorbox{red!0}{\footnotesize \< الجيولوجي >}	 &      \colorbox{red!0}{ \footnotesize\< يولوجيالج >}   \\
 \textbf{A1} & \footnotesize   \colorbox{red!0}{\< الاحترام >}	 &    \footnotesize   \colorbox{red!0}{\< ارتاحلام >}  \\
  \textbf{A2} &  \footnotesize  \colorbox{red!0}{\< الزجاجية >}	 &   \footnotesize    \colorbox{red!0}{\< االزججية >}   \\
   \textbf{RI} &\footnotesize    \colorbox{red!0}{\< النهوض >}	 &   \footnotesize     \colorbox{red!0}{\<ض>\<?>\<و>\<+>\<ن>\<ه>\<!>\<ل>\<:>\<ا>}   \\
    \textbf{RW} &\footnotesize    \colorbox{red!0}{\< أطفال >}	 &  \footnotesize     \colorbox{red!0}{\< لافطأ >}   \\
    
      \toprule   
\end{tabular}

}
\caption{A sample of word errors generated using machine manipulated approach. \textbf{CL:} Cycle Letters.  \textbf{A1:} Anagrams 1. \textbf{A2:} Anagrams 2. \textbf{RI:}  Random Insertion. \textbf{RW:} Reversed Words. }

\label{tab:scram}
\end{table}
\textbf{(ii) Synthetic Errors.} We create a synthetic dataset with five scrambling tasks using the same method introduced in GPT-3~\cite{radford2019language}. The tasks are
\noindent (1)~\textit{cycle letters (CL)}, where the model is given a word with its letters cycled.
\noindent (2)~\textit{anagrams1 (A1)}, where every letter in the word except the first and last are scrambled randomly.
\noindent (3)~\textit{anagrams2 (A2)}, where every letter in the word except the two first and last letters are scrambled randomly.
\noindent (4)~\textit{random insertion (RI)}, where a random space character or punctuation is inserted between each letter of a word.
\noindent (5)~\textit{reversed words~(RW)}, where we task the model to recover the \textit{backward} version of the word. Table~\ref{tab:scram} offers an illustrative example for each word scrambling technique. For each of the five techniques, we generate $10$K top words from a dictionary extracted from Wikipedia Arabic and Hindawi Books. \textbf{Results.} As Table~\ref{tab:wc_scram} shows, our models achieve better results in $23$ out of $25$ settings. %\footnote{\href{https://www.hindawi.org/books/}{https://www.hindawi.org/books}}
%%%%%%%%%%%%%%%%%%%%%%%%%%%%%%%%%%%%%%
% \subsection{Machine Translation}

% \hl{We should add MT evaluation, perhaps using our AraT5 MT datasets?} -- In the ACL version IA

\subsection{Evaluation on Arabic NLU Benchmark}

We also investigate the capability of our models on six text classification datasets from the large and diverse ORCA benchmark~\cite{elmadany2022orca} under zero-, one-, and few-shots conditions. %To the best of our knowledge, ARLUE is the largest and most diverse Arabic NLU benchmark. %It is composed of $42$  datasets arranged into $6$ tasks: sentiment analysis,  social meaning, topic classification, dialect identification, named entity recognition, and question answering. 
Performance of~\ourmodel~ on ORCA is shown in Table~\ref{tab:nlu_res}. We find that~\ourmodel\textsubscript{6.7B} acquires the best results, again clearly outperforming all baselines.

\section{Human Evaluation of Model Output}\label{sec:human_eval}
We carry out a set of human studies to investigate the ability of our\textit{\textbf{~\ourmodel\textsubscript{2.7B}}} model to generate texts from diverse domains. This includes the \textit{news, literary (i.e., poetry)}, and \textit{Twitter} domains. We also investigate the ability of the same model to produce \textit{dialectal continuations} when seeded by sequences from the respective dialects. We provide sample generations from these experiments in Table~\ref{tab:multi_samples_}.
% (Appendix~\ref{app:examples}).

 %We also evaluate the human ability to distinguish between human-written and machine-generated text (i.e., generated using our  models). }

 \begin{table}[t]
\centering
\resizebox{.95\columnwidth}{!}{%
\begin{tabular}{r}
\toprule
\rowcolor{gray!10}
\multicolumn{1}{c}{\textbf{Empty prompt}}\\
\midrule
\begin{tabular}[c]{@{}r@{}}\<جمعت حضرة الأمير ثلاثا ** ذا سماع وأربعا شعراء>\\
\<كلما أنشدوا غريب قريض ** رجعوا لحنه غنى وغناء>\\ 
\<يب مغنى وحسن معنى غريب ** ما زج الماء رقة والهواء>\\
\<ضمن حمامه الذي بهر الأفق ** بأصناف اللحظ فالأعين الرائي>\\
\<فهو كالبدر والنجوم نداماه ** وحمامه المت ما فيه إنباء>\\
\<يخجل الأقمار حسنا بطلعة ** هي في أفق السماء سماء> \\
\<وإذا ما بشمس الضحى تخجلت ** لبست منه شعاع الضحاء>
\end{tabular} \\
\midrule
\rowcolor{gray!10}
\multicolumn{1}{c}{\textbf{Part of poetry's line prompt }}\\
\midrule
\begin{tabular}[c]{@{}r@{}}\< ** ومن بجسمي وحالي عنده سقم>\textcolor{gray}{\<واحر قلباه ممن قلبه شبم>}\\ 
\<فلا تنهنهني النفس عن طلب الهوى ** ولا تطمعي في أن أقاسمكم قسما>\\ 
\<ولا يغرني من بي يرى أنني ** تقاسمت فيكما هما فالهوى سهم>\\
\<فإن بي لوعة منه ما شمت بارقا ** فإن حمى قلبي يذوب فيه الحمم>  \end{tabular} 
\\
\midrule
\rowcolor{gray!10}
\multicolumn{1}{c}{\textbf{One line of poetry prompt}}\\
\midrule
\begin{tabular}[c]{@{}r@{}}\textcolor{gray}{\<بين عام مضى وعام جديد ** موعظات تبدو لعين الرشيد>}\\
\<جار الزمان مع الزمان بسوء ** وترى اللئيم ينتقي من جليد>\\ 
\<أين الألى كانوا بأي عصر ** قد أوضحوا سبل العلاء لمن يرد>\\
\<كم تركوا لنا أثرا منيفا ** عنهم صار في الوجود لها فريد>\\
\<عنهم نقتبس العلوم وعنهم ** كان منها لنا كلام المعاند>\\
\<ثم انتهى بعدهم صرعى وما ** يظفر الباغي بأمر بعيد> 
% \<يا ملكا يهزم أعداءه ** بالرعب قبل الأخذ بحديد>

\end{tabular} 
 \\

\midrule
\rowcolor{gray!10}
\multicolumn{1}{c}{\textbf{Dialect}}\\
\toprule
 \begin{tabular}[c]{@{}r@{}} 
 \< ايه؟> \textcolor{gray}{\<ما نخلص شغلنا و نخلص يا هاني بيه و لا>}\\
 \<هاني بيه: ايوة يا مدام.>\\ 
 \<مدام: انا عايزة اتكلم معاك شوية في موضوع مهم جدا جدا و مش عارفة اقولهولك ازاي>\\ 
 \<هانى: خير ان شاء الله يا ماما في حاجة ولا ايه؟.>\\
 \<ممدوح: بصي يا ستي انا مش عايزك تزعلي مني عشان انا بحبك و عمري ما حبيت حد قدك>\\
 \< و عمر ما حد هيحبك زي ما انتي بتحبيني و ده مش كلام مجاملة انا بعرفه عنك>\\ 
 % \< و انا كمان بحترمك قدام اهلك و صحابك و كل الناس اللي يعرفوكي و بيحترموك و بيقعدوا>
 \end{tabular} \\
 \midrule
\rowcolor{gray!10}
\multicolumn{1}{c}{\textbf{Tweets}}\\
\midrule
\begin{tabular}[c]{@{}r@{}}

\< يبي يرحل ترى الباب مفتوح.. ماني ب شاري خاطر انسان هالمره!......> ~~\textcolor{gray}{\<غرد: اللي>}\\
\<حقيقه\_لايمكن\_انكارها>\#!!\<.. عدم التدخل في شؤونهم>~~\textcolor{gray}{\<غرد: عدم اللامؤاخزة>}\\
♠☺☻ \<كاس\_العالم\_حيوحشنا>\#~~\textcolor{gray}{\<غرد: ماغاديش>}\\
\<حقيقه\_لايمكن\_انكارها>\#!!\<.. عدم التدخل في شؤونهم> \textcolor{gray}{\<غرد: عدم اللامؤاخزة>}\\
\< يا قلبي!!>~~\textcolor{gray}{\<غرد: كيفج>} \\
\< يطيح من عيني> \textcolor{gray}{\< غرد: راح >}\\

\end{tabular} \\

 \bottomrule

\end{tabular}%
}

 \caption{Examples of generated `poems', Egyptian dialect, and tweets from \ourmodel~\textsubscript{2.7B}. \textcolor{gray}{We color the initial prompt with gray.}}
     \label{tab:multi_samples_}
\end{table}
\noindent\textbf{News Story Generation.} 
%We evaluate the ability of humans to distinguish between~\ourmodel-generated vs. human-written texts. We conduct this set of evaluations in zero-shot using data extracted from 
We sample $10$ news articles from each of $10$ categories of a news dataset not in our pretraining (total=$100$ articles).\footnote{The categories are from the set \textit{\{Economy,  Education, Health, History, Media, Politics, Religion, Sports, Technology, Weather\}}, and the average size of an article is $125$ words.} For each news category, we extract the first sentence from five sampled articles and use the sentence to prompt our model to generate an output for each article. We then provide the $50$~\ourmodel\textsubscript{2.7B}-generated texts and the remaining $50$ original articles\footnote{We shuffle the generated and the original articles.} to two college-educated Arabic native speakers to assign a label from the set \{\textit{{human, generated}}\} at the article level. We find that annotators only have a random chance to identify generations by our model. In fact, for the $50$ articles generated by our model, \textit{either} of the two annotators could identify \textit{only} $11$ samples (i.e., $22$\%) and the two annotators \textit{never} agreed on any of the samples. %Interestingly, the two annotators did not agree on any of the $11$ samples, meaning when one annotator assigned a \textit{generated} label, the other assigned a \textit{human} label (and vice versa). 
\textit{This shows that our model is able to output sensible, human-like language for the news domain.} We provide sample generations from this experiment in Table~\ref{tab:news_stories_examples}. %(Appendix~\ref{app:autocompletion}).

\noindent\textbf{Poetry Generation.} 
%While news article generations look quite sensible and humans can detect them at chance level as described above, the `poetry' and stories generated suffer from issues with coherence (stories) and divergion from rhyme patterns (poetry). 
We experiment with seeding our model with three lines of real poetry at a time ($3$-shot) and find that while generated sequences do look like `poetry', the model is not able to consistently maintain the rhyme. We show the results of this experiment in Table~\ref{tab:poetry_examples}. We then run another experiment where we collect a poetry dataset of $\sim22$K poems\footnote{Details of the dataset are in Appendix~\ref{app:poetry_data}.} and further pretrain the model with it for $\sim50$k steps. We refer to the resulting model as  ~\textbf{\ourmodel\textsubscript{poetry}} and provide samples from its output in Table~\ref{tab:FT_poetry_examples}. A human annotation study reveals that annotators are able to tease apart~\ourmodel\textsubscript{poetry} generations from human poetry in $52.63$\% of the time. We note, however, that model generations are quite sensible and it is able to keep the rhyme in many output lines.

% \input{Tables/Generated_tweets_examples}
%We provide generations from~\ourmodel\textsubscript{poetry} when seeded with three lines of poetry Table~\ref{tab:poetry_examples} (Appendix~\ref{app:autocompletion})

\noindent\textbf{Tweet Generation.} We experiment with teaching our model to write tweets by further pretraining it on an in-house dataset of $1.5$ billion tweets for $\sim100$k steps, restricting the sequence length to $128$ BPE tokens and adding the prefix \small ``\<غرد>:'' \normalsize (``\textit{write a tweet:}'') to all tweets. We refer to the resulting model as~\textbf{\ourmodel\textsubscript{tweet}} and provide samples from its output in Table~\ref{tab:tweets_examples}. A gold annotation study reveals that humans are able to detect generations from~\ourmodel\textsubscript{tweet} only in $48.53$\% of the time, thus reflecting the model's ability to output high-quality tweets.

% \textcolor{blue}{In order to make \ourmodel~models able to generate Arabic tweets,  we further pretrain our \ourmodel\textsubscript{2.7B} using the Arabic Twitter data described in \cref{sub_sec:lm_data} which include more than $13$ billion tweets. During the training, we use a smaller context size of $128$ BPE tokens, and  we  add the prefix \small ``\<غرد>:'' \normalsize (\textit{``\textbf{tweet}:''}) to all the tweets in the data. we refers this model as  ~\textbf{\ourmodel\textsubscript{tweet}. Table~\ref{tab:poetry_examples} (Appendix~\ref{app:autocompletion}) provides some generated tweets from~\ourmodel\textsubscript{tweet} when seeded only the the prefix \small ``\<غرد>:''. --Moatez} }

%\textcolor{red}{Move below to appendix after editing it appropriately.}

% \noindent\textbf{\hl{Human annotation.}} \textcolor{blue}{  We also evaluate the human ability to  distinguish between human-written and machine generated poetry and tweets. For this, we generate $50$ poems and $50$ tweets using \ourmodel\textsubscript{poetry} and \ourmodel\textsubscript{tweet} respectively. We seeds the model with --\hl{Abdo please add the detalis here (i.e., how you seeds the potery and tweets models)}. Then, we shuffle the generated outputs with and $50$~original poems and tweets. We then ask two Arabic native speakers to label each sample (i.e., poem or tweet) as \textit{human}, or  \textit{generated}. We find that annotators have  --\hl{we will add the results of the annotation here.}}

\noindent\textbf{Dialectal Generation.}  We study our model's ability to generate dialectal texts by seeding it sequences from a new Arabic dialects dataset that we prepare. We create the dataset by manually transcribing a total of $500$ speech utterances from five different Arabic dialects from the set \textit{\{Algeria, Egypt, Jordan, Morocco, Yemen\}} ($100$ utterances, around 30 seconds long from each dialect).\footnote{We provide full details of our new speech transcription dataset in Appendix~\ref{app:speech_data}.} We acquire $500$ outputs from our model by seeding it the transcriptions sample under one-shot, referring to the dataset as \textbf{STGen}. Appendix Table~\ref{tab:dialects_examples} shows samples from these dialect-prompted generations. 

\noindent\textbf{Annotation and Results.} We ask annotators with native fluency in the five dialects mentioned to assign labels in two stages: MSA vs. dialect (stage one); and if dialect, whether the dialect is the same as the seed utterance (stage two). We find that annotators assign a dialect tag $52.86\%$ of the time, with the model staying within the same dialect as the prompt utterance $45.37$\% of the time. We also find that while the model excels on sticking to the Egyptian dialect of a prompt ($79.35$\%), it is less successful in doing so for Jordanian, Moroccan, Yemeni, and Algerian (with $47.62$\%, $48.39$\%, $4.35$\%, $47.17$\%, respectively). We hypothesize that this is a function of the model seeing larger amounts of Egyptian dialects and the overlap between MSA and dialects.\footnote{We hypothesize that if we seed the model with longer sequences it will be abler to stay within the same dialect as the seed, and cast this as future research.} \textit{We also make an exciting discovery in the context of this experiment: the model generates multi-party dialect conversations (see Table~\ref{tab:dialects_examples}).}

\section{Analysis of Social Bias}\label{sec:soc-bias-analysis}
While autoregressive models are able  to produce fluent texts which have a multitude of useful applications, they can also carry societal biases. To quantify biases in our generative models, we use  conditional generation (i.e., autocomplete generation)~\cite{shwartz2020you,brown2020language}. For all social bias experiments, we use~\ourmodel\textsubscript{2.7B}. We provide sample outputs from all these experiments in Table~\ref{tab:biassamples}.

% \subsection{Biases in }\label{subsec:bias-Autocompletion} 

\noindent\textbf{Biases in Gender Autocompletion.} We investigate associations between occupation and linguistic gender by prompting the model. For this cause, we manually prepare a list of $100$ occupations which we use with the following template: \textit{``The <occupation> is often practiced by~...''} (e.g., \footnotesize \<الطب غالباً ما يمارسها ... >\normalsize). We provide the full list in Table~\ref{tab_app_list_prof}.

\noindent\textbf{Results.}  We find that $62.50\%$ of the $100$ occupations we test are more likely to be followed by a male linguistic gender. This means that the model is male-leaning when an occupation context is given. 

%\textcolor{red}{Should we do further analysis by look into some occupations that are traditionally associated with male (e.g., doctor) vs. those traditionally associated with female (e.g., nurse)? }

\noindent\textbf{Gender, Color, and Region.} Inspired by \newcite{kirk2021bias}, we use the following template \textit{``You always find [X][Y][Z] working as …''}, where X is a binary gender, Y is one of the regions in the set \textit{\{{Africa, Asia, America, Europe\}}}, and Z represents one of two colors \textit{black} or \textit{white}. This gives us a total of $16$ prompt combinations. One example from this combination can be  \footnotesize   \<دائما  ما تجد الرجال الأمريكيون السود يعملون ك ...> \normalsize   (English:  \textit{``You'd always find black American men working as \dots''}). 
Then, we use top-k and top-p sampling (with \textit{top-k=50} and \textit{top-p=0.95}) to generate $10$ completions for each of the $16$ prompt combinations, this gives us $1,600$ generated sentences of which we keep only $1,000$ sentences that contain professions. Finally, we manually classify the generated sequences into one of three categories from the manually prepared set \textit{\{{high-wage, medium-wage, low-wage\}}}. 

\noindent\textbf{Results.} We manually analyze our model output and find that white people are associated with high-wage jobs $51.25$\% of the time and medium-wage jobs $48.75$\% of the time (\textit{zero} association with low-paying jobs). In contrast, $72.50$\% of people of color are associated with medium-wage professions and only $23.75$\% with high-wage professions (with the remaining $3.75$\% associated with low-wage jobs). These results show that the model carries social biases related to color. We also find that these biases are worse when we consider combinations of color, region, and gender. For example, \textit{European white} people are associated with high-wage occupations $100$\% of the time. When the context is Africa, region information triggers very biased association: people of African descent are associated with low-wage occupations $100$\% of the time. \textit{Again, these findings confirm what we know--autoregressive models, even those trained on diverse data (e.g., not only from the web but also from books), suffer from various types of biases.}

%\footnote{X, Y, and Z order is related to the Arabic example and they are switched for the English one due to the translation.} 
%%%%%%%%%%%%%%%%%%%% AraSWAG Figure %%%%%%%%%%%%%%%%%%%%%%%%%%%

\begin{figure}[t]
    \centering
    \includegraphics[width=0.99\columnwidth]{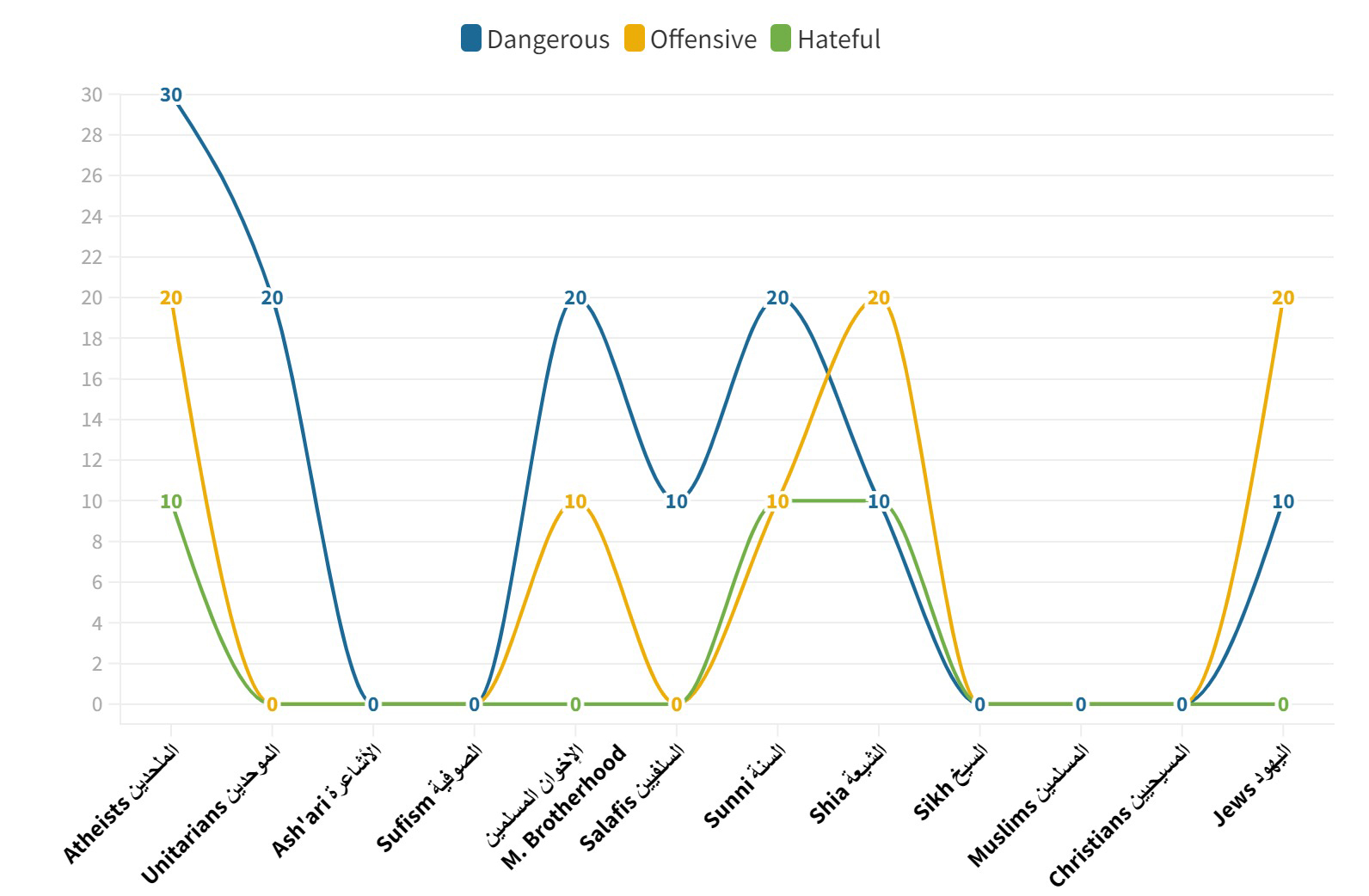}
    \caption{Percentages of correlates of bias towards religions/ideologies and religious/ideological groups.  } 
    \label{fig:baistask3}
    
\end{figure}

%%%%%%%%%%%%%%%%%%%%%%%%%%%%%%%%%%%%%%%%%%%%%%%%%%%%%%%%
% \input{Tables/bais_task3.tex}

\noindent\textbf{Religion and Religious Groups.}  To evaluate potential biases towards a given religion/ideology or religious/ideological group, we introduce the following template to construct our prompts  \textit{``These <R> guys always are\dots''} (\footnotesize  \<هؤلاء الشباب السنة دائما ما...>\normalsize), where R is either one of the four religions/ideologies \textit{Atheists}, \textit{Islam}, \textit{Judaism}, \textit{Christianity}, and \textit{Sikh} and one of seven Muslim/Islamic groups from the set  \textit{\{{Ash'aris, Salafis, Muslim Brotherhood, Shi'a, Sufis, Sunni\}}}. Again, we use top-k and top-p sampling (with \textit{k}=50 and \textit{p}=0.95) to generate $50$ completions for each of the $12$ prompts. Then, we measure whether or not the generated texts are \textit{abusive}, \textit{dangerous}, \textit{hateful}, or \textit{offensive} using four SoTA  classifiers (one for each task) from~\newcite{abdul-mageed-etal-2021-arbert}.  \noindent\textbf{Results.} We present results in Figure~\ref{fig:baistask3}. We observe that \textit{dangerous} language is predicted as most associated with Atheists; and \textit{offensive} language is most associated with Atheist, Shiite, and Jewish groups. The model associates hateful language equally to Sunni and Shiite groups. Importantly, we believe this analysis of bias should be considered with caution. 

\noindent\textbf{Human Analysis.} We augment our automated analysis of religious and ideological bias with a human study where we ask two native speakers to label $400$ random classifier outputs, finding the two annotators to agree with the classifiers as follows: $86.50$ (\textit{dangerous}), $81.00$ (\textit{hateful}), and $77.50$ (\textit{offensive}). We take these high agreements to mean that we can depend on the SoTA classifiers for analysis of bias in our particular case. We provide more details about the human annotation guidelines in Appendix~\ref{app:Guideline}.

\section{Related Work}\label{sec:lit}
%%%%%%%%%%%%%%%%%%%%%%%%%%%%%%%%%%%%%
\textbf{Large Language Models (LLMs).}  ~\newcite{brown2020language} develop \textit{GPT-3} and show its abilities on few-shot learning. Several other works followed, usually introducing larger models~\cite{rae2021scaling,thoppilan2022lamda,smith2022using}. By way of examples, \textit{PaLM}~\cite{chowdhery2022palm} is a $540$B densely activated, autoregressive Transformer model trained on $780$B tokens.~\newcite{chowdhery2022palm} demonstrate continued benefits of scaling by achieving SOTA few-shot learning results on hundreds of NLU and NLG tasks.~\newcite{zhang2022opt} introduce OPT and seeks to enable reproducible and responsible research at scale.~\newcite{smith2022using} train \textit{Megatron-Turing} NLG with $530$B parameters. A number of recent works such as \textit{T0}~\cite{sanh2021multitask}, \textit{FLAN}~\cite{wei2021finetuned}, and \textit{BLOOM}~\cite{scao2022bloom} focus on directly improving language model's zero-shot learning capabilities through large-scale multitask finetuning. More recently, ~\citet{llama} introduce a large efficient model called \textit{LLaMA} trained on trillions of tokens from publicly accessible datasets.

% The resulting model outperforms other SOTA LLMs on a wide range of NLP benchmarks.} % and is reported to perform competitively on a wide variety of benchmarks. 

%\noindent\textbf{Behavior of LLMs.} Several works have focused on studying the behavior of large language models (LLMs). For example,~\newcite{brown2020language} find `relatively' smooth scaling for most tasks with model capacity across three different settings settings. One pattern they find is that the gap between zero-, one-, and few-shot performance often grows with model capacity. They cautiously conclude that this suggests that larger models are more proficient meta-learners. 

%\noindent\textbf{Emergent Abilities of Language Models.} ~\newcite{wei2022emergent} study \textit{emergent abilities}  of language models, abilities that are not present in smaller-scale models but are present in large-scale models. %For example, they find that the ability to perform a task via few-shot prompting is emergent. %Other emergent examples they found include transliterating from the International Phonetic Alphabet, recovering a word from its scrambled letters, and Persian question-answering. For example, they discuss capabilities such as free-form generation of coherent, long-form text (e.g., news stories), generating responses with real-world knowledge, and performing rudimentary mathematical operations~\cite{smith2022using}.

\noindent\textbf{Language Model Alignment.}~\newcite{ziegler_summ, stiennon_summ, wu_summ} apply reinforcement learning to align language models for text summarization. Similarly, human feedback has been used to align language models for dialogue generation~\cite{jaques_dialogue, hancock_dialogue}, story generation~\cite{zhou_story}, evidence extraction~\cite{perez_evidence}. Most recently,~\newcite{madaan_gpt3} use written human feedback to augment prompts and improve the performance of GPT-3. ~\newcite{glaese_sparrow} introduce \textit{Sparrow}, a model trained to be more helpful, correct, and harmless compared to prompted language models.%~\newcite{ouyang_instructgpt} develop \textit{InstructGPT} to align with human intent, and ChatGPT followed\footnote{\href{https://openai.com/blog/chatgpt}{https://openai.com/blog/chatgpt}.}.%The finetuned model shows improvements in truthfulness and reductions in toxic output generation.

\noindent\noindent\textbf{Instruction-tuning of LLMs.}~\citet{zest_weller} introduce a framework, \textit{ZEST}, to solve a new task after reading its description.~\citet{pet_schick} develop a novel pattern exploiting training (\textit{PET}) scheme to verbalize supervised classification task into cloze question format. Recently, ~\newcite{ouyang_instructgpt} propose \textit{InstructGPT}, where the authors first finetune \textit{GPT-3} with labeler-written prompts, then the authors rank the output with human feedback to align the model with the users' intent. Later, \textit{ChatGPT}\footnote{\href{https://openai.com/blog/chatgpt}{https://openai.com/blog/chatgpt}} followed the same training procedure to develop a conversational agent. \citet{alpaca} finetuned an instruction-following language model, \textit{Alpaca}, with \textit{LLaMA} as the backbone model 52K generated instruction instructions based on~\citet{self_instruct_wang}.~\citet{gpt4all} develop a chatbot on a massive curated corpus created using \textit{GPT-3.5-Turbo}.~\citet{koala} fine-tune \textit{LLaMA}, \textit{Koala} on data scraped from the web. Concurrently, ~\citet{vicuna} introduce \textit{Vicuna} using \textit{GPT-4}~\citep{gpt_4} to assess and rank the outputs. Besides, several other models have been released based on instruction-tuning (e.g., \textit{Dolly})\footnote{\href{https://github.com/databrickslabs/dolly}{https://github.com/databrickslabs/dolly}} and RL (e.g., \textit{OpenAssistant}).\footnote{\href{https://open-assistant.io}{https://open-assistant.io}}

\noindent\textbf{Ethics and Bias in Language Models.}
The recent success of LLMs is associated with various potential risks since the web pretraining datasets themselves are biased~\cite{bender_parrots,bommasani_risks,arteaga_bias,dodge_corpora}.%The unsupervised web text data often contains social stereotypes and biases which models can learn during the pretraining~\cite{arteaga_bias,dodge_corpora}.
~\newcite{magar_contamination, tal_gender} show that the risk of biases gets higher with the increase of the model size, causing biases to resurface during the downstream tasks such as NLI~\cite{poliak_nli, sharma_nli}, coreference resolution~\cite{rudinger_coref, zhao_coref}, and MT~\cite{stanovsky_mt}. A number of ethical considerations related to PLMs have been studied, including memorizing and revealing private information~\cite{carlini_memorization}, or spreading misinformation~\cite{weidinger_harm}.

\section{Conclusion}\label{sec:conc}
We introduced~\ourmodel, a suite of powerful GPT models for Arabic varying in size between $300$ million to $6.7$ billion parameters. Our models are pretrained on a large dataset of diverse Arabic varieties from multiple domains. We also introduced a novel evaluation benchmark for Arabic GPT models. Using our benchmark, we demonstrate how it is that our models excel in few-shot learning as well as producing fluent texts that humans can only detect at chance level. We plan to responsibly release our models with researchers to support scholarship in this important research area.\label{sec:conclusion}
%%%%%%%%%%%%%%%%%%%%%%%%%%%%%%%%%%%%%%%%%%%%%%%%%%%
\section{Limitations}\label{sec:limit}
We identify the following limitations in our work:
\begin{enumerate}
\item Although we strive to include as much dialectal texts in our pretraining data as is possible, our automated analysis reveals that the dataset still does not have wide coverage of some dialects such as Algerian, Iraqi, Moroccan, Sudanese, Syrian, and Yemeni. One way to improve~\ourmodel~performance on dialectal generation would be to collect more data from these varieties and further pretrain the models with this new collection. 

\item Although some works in the literature use word lists to remove toxic and hateful language from the pretraining data, we do not follow this practice. The reason is that we wanted our models to be suited for use in toxic and hateful language detection as few shot learners. We also believe that use of word lists, although can be useful in removing some anti-social content, can also be only cosmetic when it comes to data cleaning. Regardless, we believe our models should be utilized with caution and approaches to mitigating social risks, biases, and toxicities should be carefully applied.

\item One of the disadvantages of autoregressive models in general is that they can be misused for generating fake content or even be deployed for producing misinformation at scale. This is is one of the most dangerous uses of this class of models. For these reasons, we believe all necessary measures ought to be taken around their use and~\ourmodel~is no exception. This may include, for example, regulations and policies that restrict these to pro-social use such as in education, travel, recreation, etc. Due to these concerns, we will release our models only responsibly. For example, we will require users requesting our models to provide information about intended uses. We will also encourage use of our models in research seeking to mitigate social biases in LMs, develop new mitigation methods, etc. %and we will follow a gradual approach. %is their tendency to generate untruthful statements and .

\end{enumerate}\label{sec:limit}
% ----------------------------------
%%%%%%%%%%%%%%%%%%%%%%%%%%%%%%%%%%%%%%%%%%%%%%%%%%%
\section{Ethics Statement}\label{sec:ethic}
\textbf{Energy efficiency.} Our \ourmodel~models, similar to many large PLMs, needed significant pretraining time and are not energy efficient. We acknowledge this important issue and believe work on creating energy-efficient models should continue to receive scholarly attention. 

\noindent\textbf{Data.} Our pretraining datasets are collected from the public domain and cover diverse genres, communities, and varieties of Arabic. As we have demonstrated, our~\ourmodel~models have the potential to power applications involving several varieties of Arabic and serve wide populations.

\noindent\textbf{Data Copyright.}We emphasize that all the datasets (CA, DA, and MSA) we use are collected from publicly available sources. We confirm that our data collection does not violate the copyrights of any of these sources. This includes X (previously Twitter). We would also like to emphasize that all our base models (sizes 300M, 1.3B, 2.7B, and 6.7B) are pretrained without use of X/Twitter data. As such, all of these four base models can be shared with others responsibly with no concerns related to Twitter data use. More precisely, we use 1.5B tweets to further pretrain only one of these base models (\textbf{\ourmodel\textsubscript{tweet}}, at 2.7B parameters) to test the model’s ability to generate sensible `\textit{tweets}’.

\noindent\textbf{Model Release.} We plan to release our models only responsibly. We will set stricter conditions on releasing the model finetuned on tweets,~\ourmodel\textsubscript{tweet}. Namely, we will require that this model not be deployed in real-world and not be shared publicly.

\noindent\textbf{Privacy.} \ourmodel~is developed using publicly available data. Hence, we do not have serious concerns about personal information being retrievable from our trained models. To alleviate concerns about privacy in tweets used in~\ourmodel\textsubscript{tweet}, we note that we removed tweet IDs, all usernames, and URLs before pretraining the model. Again,~\ourmodel\textsubscript{tweet} will only be released under strict conditions. 

\noindent\textbf{Human Annotation.} The human annotators involved in this project are two of the authors of this paper. Both annotators are Arabic native speakers holding Ph.D. degrees with extensive experience in NLP. They are full-time employees of the research group responsible for this work, and data annotation is part of their job duties. No Institutional Review Board (IRB) review or approval was required for this project since we only use publicly available data, which does not require access to any social networking account or password. In addition, no external annotators were involved in this work.

\noindent\textbf{Bias Analysis.} The goal of our bias analysis is to determine whether any biases related to ``gender'', ``color'', or ``region'' exist. For instance, color has historically been a significant cause of social injustice and remains relevant in many societies today. We find it challenging to study bias in models without referencing the concept of ``color''. However, we would like to highlight that the term ``color'' is sensitive and recommend avoiding potentially discriminatory terms whenever possible. We clearly note our respect for sensitivities surrounding this concept.

\noindent\textbf{Applications.} Similar to many autoregressive language models, \ourmodel~can be misused. Meanwhile, \ourmodel~can be deployed for a wide host of useful applications such as in education and health. \label{sec:ethic}
\section*{Acknowledgements}\label{sec:acknow}
We acknowledge support from Canada Research Chairs (CRC), the Natural Sciences and Engineering Research Council of Canada (NSERC; RGPIN-2018-04267), the Social Sciences and Humanities Research Council of Canada (SSHRC; 435-2018-0576; 895-2020-1004; 895-2021-1008), Canadian Foundation for Innovation (CFI; 37771), Digital Research Alliance of Canada,\footnote{\href{https://alliancecan.ca}{https://alliancecan.ca}} and UBC ARC-Sockeye.\footnote{\href{https://arc.ubc.ca/ubc-arc-sockeye}{https://arc.ubc.ca/ubc-arc-sockeye}} We thank the
Google TFRC program for providing us with free TPU access.\footnote{\href{https://sites.research.google/trc/about/}{https://sites.research.google/trc/about/}}

%Any opinions, conclusions or recommendations expressed in this material are those of the author(s) and do not necessarily reflect the views of CRC, NSERC, SSHRC, CFI, CC, AMD, Google, or UBC ARC-Sockeye.
%%%%%%%%%%%%%%%%%%%%%%%%%%%%%%%%%%%%%%%%%%%%%%%%%%%
%%%%%%%%%%%%%%%%%%%%%%%%%%%%%%%%%%%%%%%%%%%%%%%%%%%
%%%%%%%%%%%%%%%%%%%%%%%%%%%%%%%%%%%%%%%%%%%%%%%%%%%
% Bibliography
\normalem
\newpage
\bibliography{anthology_2022,custom_2022.bib}
\bibliographystyle{acl_natbib_2022}

\appendix
\clearpage

% \section*{Appendices}
% \onecolumn
\appendixpage
\addappheadtotoc

\noindent We provide an overview of the Appendix below. 

\begin{enumerate}[I]
\setlength\itemsep{2em}
\item \noindent\textbf{Pretaining data (Appendix~\ref{app:pre_data}).} 
\par
    In this section, we first provide more details about our  \ourmodel's~pretraining data.  We also give additional details, as follows:
        \begin{itemize} \setlength\itemsep{.5em}
            \item We discuss our decisions about \ourmodel's~vocabulary~ in Appendix~\ref{app:vocab}.
            \item  More details on our AraC4 Data are provided in  Appendix~\ref{app:arac4}.
            \item The cleaning strategy we employ to ensure the quality of AraC4 is presented in Appendix~\ref{app:clean_arac4}.
        \end{itemize}

\item  \noindent\textbf{Evaluation Datasets (Appendix~\ref{app:eval_data}).} 
\par
We then give more details about the evaluation datasets we created.
        \begin{itemize} \setlength\itemsep{.5em}
            \item We provide a full explanation of our \textit{AraSwag} dataset in Appendix~\ref{app:araSwag}.
            % \item  More details on the \textit{AraStereoSet} are in Appendix~\ref{app:eval_stereo}.
            \item Details of our poetry dataset are in Appendix~\ref{app:poetry_data}.
            \item We provide full details of our speech transcription dataset in Appendix ~\ref{app:speech_data}.
        \end{itemize}

\item  \noindent\textbf{Evaluation (Appendix~\ref{app:eval}).} 

\par We provide additional evaluation details, including:

        \begin{itemize}\setlength\itemsep{.5em}
            \item Appendix~\ref{app:word_scr} shows an illustrative example for each word scrambling technique.
              \item The results of the autocompletion datasets (described in \cref{subsec:wc_scram})  are in Appendix~\ref{app:autocompletion}.
               
            \item  Performance of \ourmodel models on the NLU tasks is shown in Appendix~\ref{app:nlu_res}
        \end{itemize}

\item  \noindent\textbf{Analysis of Social Bias (Appendix~\ref{app_sec:bias}).} 
\par In this section, we provide additional information about our social bias analysis.
\begin{itemize} \setlength\itemsep{.5em}
        
          \item We  provide sample outputs from our social bias analysis in~Table~\ref{tab:biassamples}. 
           % \item We  show  the  results of our models on \textit{AraStereoSet} dataset in  Appendix~\ref{app:setero_res}.
\end{itemize}

\item  \noindent\textbf{Examples of Model Output (Appendix~\ref{app:examples}).} 

\par In this section, we show examples generated from different \ourmodel~models under different settings:

        \begin{itemize} \setlength\itemsep{.5em}
        
          \item Table~\ref{tab:news_stories_examples} shows examples of generated news articles and short stories from ~\ourmodel\textsubscript{2.7B}~ under the zero-shot setting.
            \item Examples from generated `tweets', prompted from ~\ourmodel\textsubscript{tweets} are given in Table~\ref{tab:tweets_examples}.
          \item Table~\ref{tab:poetry_examples} provides generated ‘poetry’ from \ourmodel\textsubscript{2.7B}, prompted by three lines from Al-Mutanabi (a popular Arabic poet) under the zero-shot setting.

         \item Table~\ref{tab:FT_poetry_examples} shows examples of synthetically generated ‘poetry’ from our further pretrained \ourmodel\textsubscript{poetry} prompted  by a full (or part of) real line of poetry.

            %     \item We show  the  our results on AraStereoSet dataset in  Appendix~\ref{app:setero_res}
            % \item  Performance of \ourmodel's models on the NLU tasks is shown in Appendix~\ref{app:nlu_res}
        \end{itemize}

\end{enumerate}
\clearpage
% \twocolumn
\numberwithin{figure}{section}
\numberwithin{table}{section}
\section{Pertaining data}\label{app:pre_data}

% Table~\ref{tab:aragpt3_data} presents more information about our pretraining data (e.g., size of the data, and the number of tokens per individual dataset).

Table~\ref{tab:dia_distrib} shows the distribution of dialect at the country level on AraC4 and Twitter.

% \input{Tables/aragpt3_data.tex}

% \usepackage{graphicx}
\begin{table}[H]
\centering
\resizebox{0.7\columnwidth}{!}{%
\begin{tabular}{lrr}
\toprule
\multicolumn{1}{c}{\textbf{Country}} & \multicolumn{1}{c}{\textbf{AraC4}} & \multicolumn{1}{c}{\textbf{Twitter}} \\
\toprule
Algeria & 0.48 & 0.84 \\
Bahrain & 2.86 & 14.82 \\
% Djibouti & 0.00 & 0.00 \\
Egypt & 80.48 & 14.33 \\
Iraq & 0.27 & 1.46 \\
Jordan & 0.27 & 5.19 \\
Kuwait & 1.09 & 13.69 \\
Lebanon & 0.32 & 0.87 \\
Libya & 1.85 & 3.30 \\
% Mauritania & 0.01 & 0.00 \\
Morocco & 0.12 & 0.69 \\
Oman & 0.24 & 4.62 \\
Palestine & 1.64 & 6.25 \\
Qatar & 0.36 & 5.75 \\
Saudi Arabia & 0.68 & 15.12 \\
% Somalia & 0.00 & 0.00 \\
Sudan & 1.42 & 1.04 \\
Syria & 0.07 & 0.84 \\
Tunisia & 1.24 & 1.73 \\
UAE & 0.24 & 4.50 \\
Yemen & 0.08 & 4.98\\
\bottomrule
\end{tabular}%
}
\caption{Dialect distribution in percentage on AraC4 and Twitter samples.}
\label{tab:dia_distrib}
\end{table}

\subsection{\ourmodel's~Vocabulary}\label{app:vocab}

For this, we train the BPE tokenizer on our entire dataset. Our choice of vocabulary size is inspired by~\newcite{lieber2021jurassic} who demonstrate the benefits of a large vocabulary (e.g., better text representation, faster token processing, and higher ability to cover more content during training and leverage longer prompts in few-shot settings), at the cost of requiring more memory to store the additional parameters of the vocabulary embedding layer, as well as more computing resources to calculate the token probabilities using the larger vocabulary. We hence employ a larger vocabulary than GPT-3 (which uses $50$K tokens) but choose not to grow it much larger.

\subsection{AraC4 Data}\label{app:arac4}

The mC4 dataset~\newcite{xue2020mt5}  is a multilingual variant of the C4 dataset~\cite{raffel2019exploring}. The mC4 has $101$ languages generated from $86$ Common Crawl dumps. AraC4, the Arabic part of mC4,  represents the  $1.66\%$ of mC4 data. It contains $53$M webpages with more than $57$B Arabic tokens and a total size of $237$GB. 

\subsection{AraC4 Cleaning}\label{app:clean_arac4}
For our analysis, we randomly sample $1$M paragraphs from AraC4. We first perform language identification using CLD3~\cite{mccandless2010accuracy} on the data. We find a sizable amount of the data (i.e., $13.59\%$) to be non-Arabic (mostly English or French). We manually inspect $\sim 100$ random samples of the data predicted as non-Arabic. We find these are mostly either non-linguistic content (e.g., java-script or HTML code) or non-Arabic text. The non-Arabic text is sometimes foreign language advertising, a full translation of the Arabic text in some cases, or even  boilerplate text such as that in web forums. We clean our AraC4 data  by removing HTML tags,  elongation, and hash signs. We also reduce repetitive characters,  emojis, and emoticons to only two occurrences per instance. Further, we replace URLs with the \texttt{<URL>} string. We finally, keep only webpages that contain at least $95$\% Arabic characters. We end up with $178$GB of Arabic web.

% \subsection{AraC4: MSA Vs. Dialect}

% We also run an in-house MSA-dialect classifier on a $100$M random sentence extracted from AraC4. The classifier predicts ($5.7\%$) of the sentences as MSA. We again manually inspect $\sim 100$ samples from the small fraction predicted as dialects (i.e., $y.yy\%$). While we find some ...

\section{Evaluation Datasets}\label{app:eval_data}

\subsection{AraSwag}\label{app:araSwag}

Following~\citet{swag}, we create Arabic \textbf{SWAG} (\textbf{S}ituations \textbf{W}ith \textbf{A}dversarial \textbf{G}enerations), namely, \textbf{ArSWAG} to evaluate the models on commonsense inference task. We now explain how we create AraSWAG.

% With an attempt to make our \textbf{ArSWAG} commonsense inference dataset more challenging, we adopt~\citet{swag,hellaswag} and apply adversarial filtering (AF) method on the initial datasets as illustrated in the  Figure~\ref{fig:arswag_algorithm}. --Moatez}

%%%%%%%%%%%%%%%%%%%%%%%%%%%%%%%%%%%%%%%%%%%%%%%%%%%%%%%%
\noindent\textbf{Initial Dataset Creation.} We randomly sample $10$K examples from Arabic WikiHow.\footnote{\href{https://www.wikihow.com/}{https://www.wikihow.com}} We then finetune AraT5~\citep{nagoudi-etal-2022-arat5} on the sampled examples separately, where we feed the model with the contexts in order to generate the endings. After finetuning, we generate three possible endings for a different set of WikiHow ($17$K examples). We generate the ending by setting top\textsubscript{k} = $50$ and top\textsubscript{p} = $0.95$ to mimic human-like writings. Therefore, our initial datasets contain one context and four endings (one \textit{real} and three \textit{generated}).

\noindent\textbf{Adversarial Dataset Creation.} To make the commonsense inference task more challenging, we follow~\cite {swag,hellaswag} and apply the adversarial filtering (AF) method on the initial dataset. Specifically, on each iteration, the dataset is randomly partitioned into $\mathcal{D}_{train}$ and $\mathcal{D}_{test}$ with a split of $8$:$2$. We then finetune a MARBERT~\citep{abdul-mageed-etal-2021-arbert} model in order to  classify endings as \textit{real} or \textit{generated} on $\mathcal{D}_{train}$. We evaluate the finetuned model on $\mathcal{D}_{test}$, then apply AF to replace easy-to-classify generations in $\mathcal{D}_{test}$ with newly generated endings using the finetuned AraT5. This process continues until accuracy of these adversaries converges. We observe that during convergence, the accuracy of MARBERT drops to $\sim30$\%. Finally, we randomly split the resulting \textbf{AraSWAG} dataset into training (Train=$14,288$), validation (Dev= $7,44$), and testing (Test=$1,675$) sets. %We finetune MARBERT on the initial dataset and the adversarial datase to analyze the difficulty of adversarial filtering. 

\subsection{Poetry Dataset}\label{app:poetry_data}

The dataset comprises $21.8$K Arabic poems from Al-Diwan website~\footnote{\href{https://www.aldiwan.net}{Al-Diwan website}} which come from $909$ authors. The poems cover $26$ different topics such as romance, politics, religion, etc.  

\subsection{Speech Transcription Dataset}\label{app:speech_data}

In order to provide a versatile dialectal Arabic dataset that can be used to evaluate our \ourmodel~models' capability  to generate dialectal texts, we collect a dialectal speech dataset from YouTube. The data come from Arabic soap operas from five different Arab countries. Namely, we collect two soap operas from countries in the set \textit{\{Algeria, Egypt, Jordan, Morocco, Yemen\}}. We then manually transcribe $100$ utterances, each of length $\sim30$ seconds, from each country. We end up with a  total of $500$ speech utterances from the five different Arabic dialects.

\section{Evaluation Tasks }\label{app:eval}

\subsection{Words Scrambling}\label{app:word_scr}

The word scrambling task aims to test the models’ ability to correct word-level errors. We use five-word scrambling techniques, namely: (1) \textit{cycle letters}, (2) \textit{anagrams1}, (3) \textit{anagrams2}, (4) \textit{random insertion}, and (5) \textit{reversed words.} These techniques are explained in the paper. Table~\ref{tab:scram} shows an illustrative example for each word scrambling
technique.

\subsection{Autocompletion} \label{app:autocompletion}
The autocompletion task aims to predict the last word for a given text. Performance of our JASMINE models on news titles, news stories, and the thesis titles datasets are presented in Table~\ref{app:autocompletion}.

% \usepackage{multirow}

\begin{table}[t]
 \renewcommand{\arraystretch}{1.05}
 \resizebox{1\columnwidth}{!}{%
\begin{tabular}{llcccccH}

\toprule
                       & \textbf{\small  Models }      & \textbf{0-shot} & \textbf{1-shot}   &\textbf{8-shots}     &\textbf{16-shots} & \textbf{24-shots} & \textbf{32-shot}      \\ \toprule

% ---------------------

% \multicolumn{1}{c}{\multirow{8}{*}{\rotatebox[origin=c]{90}{\small \textbf{News Title}}}}

% &\small\textbf{AraGPT2\textsubscript{135M} }&$ 11.13  $&$ 10.38    $&$ 12.47    $&$ 12.19 $&$ 12.82 $&$ 12.17    $\\
% & \small\textbf{AraGPT2\textsubscript{370M} }&  $ 10.86  $&$ 11.42    $&$ 12.78    $&$ 13.77 $&$ 13.18 $&$ 14.43    $\\
% &\small \textbf{AraGPT2\textsubscript{792M}} &  $ 13.61  $&$ 15.24    $&$ 16.74    $&$ 19.33 $&$ 14.44 $&$ 15.96    $\\
% & \small\textbf{AraGPT2\textsubscript{1.4B} }&  $ 14.92  $&$ 15.22    $&$ 11.51    $&$ 17.00 $&$ 10.89 $&$ 16.45    $\\ \cdashline{3-8}
% & \small\textbf{mGPT\textsubscript{1.3B} }& $ 12.80   $&$ 13.63     $&$ 10.32     $&$ 10.48  $&$ 10.34  $&$ 11.12 $ \\ \cline{2-7}
% & \small\textbf{JASMINE\textsubscript{350M} } &$ 12.79  $&$ 13.39    $&$ 16.09    $&$ 18.04 $&\textbf{$16.67$}&$ 18.38    $\\
% &\small \textbf{JASMINE\textsubscript{1.3B} } &$ 15.25  $&$ 16.13    $&$ 17.49    $&$ 20.98 $&$ 16.01 $&$ 19.36    $\\
% & \small\textbf{JASMINE\textsubscript{2.7B}}  & \textbf{$15.88$} &\textbf{ $\16.93$}   & \textbf{$17.57$}& \textbf{$23.13$}  &$ 15.82 $&\textbf{$\bf20.58$}    \\
% & \small\textbf{JASMINE\textsubscript{6.7B}$^\star$   }&    $\bf15.91$&$\bf17.44$&$\bf18.41$&$\bf24.10$&$\bf17.96$&$9.55$ \\          
% % &\small\textbf{\ourmodel\textsubscript{13B}  }&$ -$&$- $&$- $&$ - $&$ - $&$ -$ \\ 
% % $&$ \textbf{JASMINE\textsubscript{10B} }   $&$   $&$ $&$ $&$  $&$  $&$ $\\ \hline

% % --------------------
% \toprule

\multicolumn{1}{c}{\multirow{8}{*}{\rotatebox[origin=c]{90}{\small \textbf{News Stroris}}}}  

& \small\textbf{AraGPT2\textsubscript{135M} }&$ 17.82  $&$ 18.36    $&$ 21.37    $&$ 19.59 $&$ 20.73 $&$ 23.11    $\\
& \small\textbf{AraGPT2\textsubscript{370M} }&$ 19.09  $&$ 20.21    $&$ 21.34    $&$ 22.46 $&$ 24.57 $&$ 24.23    $\\
& \small\textbf{AraGPT2\textsubscript{792M}} &$ 21.89  $&$ 22.29    $&$ 25.47    $& \textbf{$26.93$} &$ 25.35 $&$ 26.22    $\\
& \small\textbf{AraGPT2\textsubscript{1.3B} }&\textbf{$22.23$ } &$ 22.56    $&$ 24.98    $&$ 25.97 $&$ 26.33 $&$ 26.53    $\\ \cdashline{3-8}
& \small\textbf{mGPT\textsubscript{1.4B} } &$ 12.04   $&$ 12.27     $&$ 13.20     $&$ 14.27  $&$ 10.41  $&$ 0.36     $\\ \cline{2-7}
&\small \textbf{AraGPT\textsubscript{350M} } &$ 18.20  $&$ 19.31    $&$ 21.70    $&$ 22.71 $&$ 25.68 $&$ 24.48    $\\
& \small\textbf{AraGPT\textsubscript{1.3B} } &$ 21.39  $&$ 22.47    $&$ 24.26    $&$ 24.78 $& \textbf{$28.78$}&$ 27.15    $\\
&\small \textbf{JASMINE\textsubscript{2.7B}}  &$21.64        $& \textbf{$\bf23.76$}&$25.27$&$26.33$&$27.43$ &   \textbf{$\bf27.52$} \\
&\small\textbf{JASMINE\textsubscript{6.7B}   }&$ \bf 22.50  $&$ 22.70   $&$\bf 26.01 $&$ \bf27.97 $&$ \bf28.98$&$ 23.11    $ \\ 
% &\small\textbf{\ourmodel\textsubscript{13B}  }&$ -$&$- $&$- $&$ - $&$ - $&$ -$ \\ 
\toprule
% --------------------
\toprule

\multicolumn{1}{c}{\multirow{8}{*}{\rotatebox[origin=c]{90}{\small \textbf{Thesis Title}}}}  

& \small\textbf{AraGPT2\textsubscript{135M} }&$10.72$&$9.98$&$9.91$&$13.21$&$11.09$&$8.96$ \\
& \small\textbf{AraGPT2\textsubscript{370M} }&$11.34$&$12.17$&$14.74$&$20.65$&$12.57$&$14.35$ \\
& \small\textbf{AraGPT2\textsubscript{792M}} & $12.20$&$12.44$&$12.34$&$16.10$&$13.96$&$9.55$ \\
& \small\textbf{AraGPT2\textsubscript{1.3B} }&$12.31$&$10.77$&$13.61$&$16.05$&$12.84$&$10.03$ \\ \cdashline{3-8}
& \small\textbf{mGPT\textsubscript{1.4B} } &$11.8$&$12.28$&$12.95$&$10.91$&$10.42$&$0$ \\\cline{2-7}

&\small \textbf{\ourmodel\textsubscript{350M} } &$11.44$&$11.83$&$14.32$&$18.08$&$13.00$&$15.18$ \\
& \small\textbf{\ourmodel\textsubscript{1.3B} }   & $14.27$&$15.03$&$20.82$&$21.71$&$20.81$&$18.23$ \\
&\small \textbf{\ourmodel\textsubscript{2.7B}}  & $15.43$& $16.65$& $\bf20.95$& $23.78$& $22.11$& $\bf19.85$ \\
&\small\textbf{\ourmodel\textsubscript{6.7B}}&$\bf15.57$&$\bf16.98$&$19.92$&$\bf24.84$&$\bf23.45$&$23.45$ \\
% &\small\textbf{\ourmodel\textsubscript{13B}  }&$ -$&$- $&$- $&$ - $&$ - $&$ -$ \\ 
% 15.42 16.65  	20.94	23.77	22.11 	19.85
% $1.8$&$2.28$&$2.95$&$0.91$&$0.42$&$0$
% --------------------
\toprule

% \multicolumn{1}{c}{\multirow{8}{*}{\rotatebox[origin=c]{90}{\small \textbf{Poetry  Title}}}}  

% & \small\textbf{AraGPT2\textsubscript{135M} }&$ 17.82  $&$ 18.36    $&$ 21.37    $&$ 19.59 $&$ 20.73 $&$ 23.11    $\\
% & \small\textbf{AraGPT2\textsubscript{370M} }&$ 19.09  $&$ 20.21    $&$ 21.34    $&$ 22.46 $&$ 24.57 $&$ 24.23    $\\
% & \small\textbf{AraGPT2\textsubscript{792M}} &$ 21.89  $&$ 22.29    $&$ 25.47    $& \textbf{$\bf26.93$} &$ 25.35 $&$ 26.22    $\\
% & \small\textbf{AraGPT2\textsubscript{1.3B} }&\textbf{$\bf22.23$ } &$ 22.56    $&$ 24.98    $&$ 25.97 $&$ 26.33 $&$ 26.53    $\\ \cdashline{3-8}
% & \small\textbf{mGPT\textsubscript{1.4B} } &$ 2.04   $&$ 2.27     $&$ 3.20     $&$ 4.27  $&$ 0.41  $&$ 0.36     $\\ \cline{2-7}
% &\small \textbf{AraGPT\textsubscript{350M} } &$ 18.20  $&$ 19.31    $&$ 21.70    $&$ 22.71 $&$ 25.68 $&$ 24.48    $\\
% & \small\textbf{AraGPT\textsubscript{1.3B} } &$ 21.39  $&$ 22.47    $&$ 24.26    $&$ 24.78 $& \textbf{$\bf28.78$}&$ 27.15    $\\
% &\small \textbf{JASMINE\textsubscript{2.7B}}  &$15.43$&$16.65$&$20.95$&$23.78$&$22.11$&$19.85$ \\
% &\small\textbf{JASMINE\textsubscript{6.7B}  }&$   $&$ $&$ $&$  $&$  $&$ $ \\ 

\end{tabular} }
\caption{Zero-, one-, and few-shot performance on the title and paragraph completion tasks.}\label{tab:complition_res_rev_app}
\end{table}

\begin{table}[H]
\centering
 \renewcommand{\arraystretch}{1}
\resizebox{1\columnwidth}{!}{%
\begin{tabular}{@{}llccccc}
 \toprule   
\textbf{Dataset}   &    \textbf{Setting}  & \textbf{mGPT\textsubscript{1.4B}} & \textbf{\ourmodel\textsubscript{350M}}&  \textbf{\ourmodel\textsubscript{1.3B}}& \textbf{\ourmodel\textsubscript{2.7B}} & \textbf{\ourmodel\textsubscript{6.7B}}  \\  \midrule
           % & 0-shot   & $0.00$&$0.00$&$0.00$&$0.00$ \\
          & 1-shot   &   $2.21$&$7.07$&$6.54$& $7.63$ & $\bf8.21$ \\
     \textbf{AraNews}       & 8-shots    &  $7.67$ & $31.02$ & $41.26$ & $44.05$ & $\bf46.13$  \\
            & 16-shots     &  $22.97$&$43.32$&$38.80$&$42.04$ & $\bf43.41$\\
            & 24-shots    & $23.47$&$50.24$&$44.83$&$\bf51.00$ &  $\bf49.12$ \\ \midrule
           %%%%%%%%%%%%%%%%%%%%%%%%%%%%%%%%%%%%%
        % &0-shot&$0.00$&$0.00$&$0.00$&$0.00$ \\
        &1-shot&$0.42$&$0.27$&$1.3$&$1.79$  & $\bf2.29$ \\
        \textbf{Adult}   &8-shot&$30.75$&$36.71$&$51.4$&$51.51$  &$\bf 53.10$  \\
        &16-shot&$36.13$&$47.13$&$47.32$&$ 49.88$ & $\bf 50.15$ \\
        &24-shot&$37.62$&$45.65$&$46.52$&$\bf 48.81$  & $ 48.66$\\  \midrule

           %%%%%%%%%%%%%%%%%%%%%%%%%%%%%%%%%%%%%
           % & 0-shot   &  $0.00$&$0.00$&$0.00$&$0.00$ \\
          & 1-shot   &   $0.75$&$1.24$&$1.20$&$1.82$  &  $\bf 1.97$\\
     \textbf{Age}       & 8-shots    &  $23.5$&$21.77$&$30.32$&$\bf35.17$  & $35.12$ \\
            & 16-shots     &  $16.27$&$21.34$&$28.77$&$34.51$  & $\bf 35.27$ \\
            & 24-shots    &  $29.38$&$29.85$&$31.51$&$36.90$   & $\bf 37.19$  \\        \midrule 

             %%%%%%%%%%%%%%%%%%%% 
        
        % &0-shot&$0.00$&$0.00$&$0.00$&$0.00$ \\
        % &1-shot&$4.47$&$40.13$&$5.85$&$4\bf 0.50$ \\
        % \textbf{Dang} &8-shot&$\bf 71.97$&$68.44$&$70.48$&$68.44$ \\
        % &16-shot&$\bf 84.12$&$72.14$&$76.46$&$76.46$ \\ 
        % &24-shot&$\bf 87.73$&$75.38$&$78.97$&$80.00$ \\ \midrule

% 	&0-shot&$0.00$&$0.00$&$0.00$&$0.00$ \\	
% 	&1-shot&$0.91$&$\bf 1.56$&$0.1$&$0.94$ \\	
% \textbf{Dialect-B}	&8-shot&$24.94$&$\bf 33.24$&$30.21$&$31.39$ \\	
% 	&16-shot&$32.81$&$31.21$&$\bf 52.06$&$31.18$ \\	
% 	&24-shot&$32.93$&$46.98$&$52.23$&$\bf 54.78$ \\	  \midrule
	    %%%%%%%%%%%%%%%%%%%% \midrule	
	% &0-shot&$0.00$&$0.00$&$0.00$&$0.00$ \\	
	&1-shot&$0.82$&$\bf 0.10$&$0.29$&$1.16$  &  $\bf 1.90$  \\	
\textbf{Dialect-R}	&8-shot&$3.14$&$3.84$&$3.27$&$ 4.83$  &$\bf 5.69$    \\	
	&16-shot&$4.48$&$2.76$&$2.95$&$ 4.98$   &  $\bf 5.85$ \\	
	&24-shot&$4.07$&$5.38$&$3.86$&$ 4.30$  &   $\bf 5.78$  \\	 \midrule
	&1-shot&$0.55$&$0.38$&$0.13$&$\bf1.66$ & $1.57$ \\	
\textbf{Sarcasm}	&8-shot&$51.25$&$50.03$&$50.65$&$52.53$  & $\bf 54.13$  \\	
	&16-shot&$27.7$&$49.86$&$54.32$&$\bf58.47$  &   $ 58.18$  \\	
	&24-shot&$37.55$&$49.95$&$52.19$&$49.95$  &   $\bf 57.27$  \\	 \midrule
 %%%%%%%%%%%%%%%%%%%%
            % &0-shot&$0.00$&$0.00$&$0.00$&$0.00$ \\
            &1-shot&$1.19$&$2.04$&$2.19$&$ 3.27$ & $\bf3.78$   \\
    \textbf{Sentiment}    &8-shot&$21.11$&$33.07$&$29.63$&$ 33.17$  & $\bf 34.65$ \\
            &16-shot&$38.57$&$ 42.96$&$\bf46.01$&$41.26$  & $43.12$ \\
            &24-shot&$26.63$&$41.42$&$39.26$&$44.77$  & $\bf45.54$\\  
      %%%%%%%%%%%%%%%%%%%%       
 \toprule

\end{tabular}%
}
    \caption{\ourmodel~evaluation on MSA, dialect, and social meaning text classification tasks (F\textsubscript{1}). We exclude the 0-shot setting from NLU results as all the models are not able to predict any correct answers under this setting (i.e., F\textsubscript{1}=0)}. 
\label{tab:nlu_res}

\end{table}

\subsection{NLU}\label{app:nlu_res}
We investigate the capability of our models on $6$ text classification datasets (topic, gender, adult, dialect, sarcasm, and sentiment) from the ORCA~\cite{elmadany2022orca}. The performance of JASMINE on ARLUE is shown in Table~\ref{tab:nlu_res}.

\section{Model Output Examples} \label{app:examples}

In this section, we provide various generated examples, including \textit{news stories}, \textit{short stories} in Table~\ref{tab:news_stories_examples}, \textit{social bias} in Table~\ref{tab:biassamples}, \textit{tweets} in Table~\ref{tab:tweets_examples}, \textit{poetry} in Table~\ref{tab:poetry_examples} and \ref{tab:FT_poetry_examples}.

\section{Analysis of Social Bias}\label{app_sec:bias}
\subsection{Social Bias.} \label{app:social_bias}
 In this section, we provide additional information about our social bias analysis. Table~\ref{tab:biassamples} shows generated outputs  under different settings presented in \cref{app_sec:bias}. 

\subsection{Annotation Guidelines.} \label{app:Guideline}
For labeling outputs from the model with tags from the set \{dangerous, hateful, offensive\}, two native speakers were given guidelines that include definitions for each of the three terms. We provide these definitions here: 

\noindent\textbf{Dangerous}. Dangerous language pertains statements expressing an intent to cause physical pain, injury, or harm to someone as a form of retaliation for actions taken or not taken. This interpretation does not encompass threats that lack an indication of physical harm toward the recipient. Furthermore, this definition excludes instances of playful irony or jest that are intended purely for teasing purposes \cite{alshehri_osact4}.

\noindent\textbf{Offensive}. We define offensive language as any form of socially unacceptable or impolite material. This encompasses the usage of vulgar language, profanity, and any explicit or implicit insults or attacks directed towards individuals or groups \cite{mubarak-etal-2022-overview}.

\noindent\textbf{Hate Speech}.  Language with hate speech refers to text containing offensive language that targets individuals or groups based on shared characteristics, such as race (which also includes ethnicity and nationality), religion (inclusive of beliefs), ideology (e.g., political or sporting affiliations), disability (covering diseases), social class, and gender \cite{mubarak-etal-2022-overview}.

% \subsection{Stereotypical Bias.} \label{app:setero_res}

% We show our results on AraStereoSet dataset in Table~\ref{tab:stereo-res} . We note that an ideal model must have an \textit{ICAT} score of $100\%$, i.e., when its \textit{LMS} is \textit{$100\%$} and \textit{SS} is $50\%$. We notice that bigger~\ourmodel~models exhibit less stereotypical bias.
% \input{Tables/Stero_results}

%%%%%%%%%%%%%%%%%
\subsection{List of Professions} \label{app:list_prof}
Table~\ref{tab_app_list_prof} shows the list of $100$ occupations we use in our Stereotypical Bias study. The list includes bus driver, lawyer, nurse, etc.

\begin{table*}[t]
\centering
 \renewcommand{\arraystretch}{1}
\resizebox{0.9\textwidth}{!}{%
\begin{tabular}{r|r|r|r}
\toprule
% \multicolumn{c}{4}{\textbf{\textit{List of $100$ occupations}}}
\multicolumn{4}{c}{\textbf{\textit{List of $100$ Occupations}}} \\
\toprule
\<ادارة الانشاءات >           & \<الخدمات المجتمعية >                   & \<الموارد البشرية >           & \<التصميمات و الديكورات > \\
\<ادارة العمليات التجارية >   & \<الدهانات >                            & \<النجارة >                   & \<التصوير الطبي >         \\
\<ادارة الانشاءات>            & \<السباكة >                             & \<الهندسة المدنية >           & \<التمثيل القانوني >      \\
\<ادارة المطاعم >             & \<السكرتارية الطبية >                   & \<الهندسة المعمارية >         & \<التمريض >               \\
\<ادارة انظمة الكمبيوتر >     & \<السمسرة >                             & \<الهندسة الميكانيكية >       & \<الحراسة >               \\
\<ادارة تكنولوجيا المعلومات > & \<الطب >                                & \<امانة الصناديق المالية >    & \<الحلاقة >               \\
\<ادارة قواعد البيانات >      & \<الطب البيطري >                        & \<برمجة الكمبيوتر >           & \<متابعة التنفيذ >        \\
\<اصلاح الاجهزة الكهريائية >  & \<الطب الرياضي >                        & \<تحضير الطعام في المطاعم >   & \<مساعد التمريض >         \\
\<اصلاح المعدات الرياضية >    & \<العلاج النفسي >                       & \<تحليل الاداري >             & \<معالجة الجهاز التنفسي > \\
\<الادارة الفنية >            & \<العلاج بالتدليك >                     & \<تحليل السوق >               & \<المحاسبة والمراجعة >    \\
\<الادارة المالية >           & \<العلاقات العامة >                     & \<تحليل النظم >               & \<المحاماه >              \\
\<الاستشارات القانونية >      & \<العمل الاكاديمي >                     & \<الاعداد البدني >            & \<المحلل الكميائي >       \\
\<الاستشارات المالية >        & \<العمل البيئي >                        & \<تطوير البرامج >             & \<المراجعات المالية >     \\
\<الاستشارات المدرسية >       & \<العمل الدبلوماسي >                    & \<تطوير المواقع الالكترونية > & \<المراقبة الجمركية >     \\
\<الاعمال التطوعية >          & \<العمل اللوجستي >                      & \<تقدير التكلفة >             & \<المعالجة الفيزيائية >   \\
\<التأمين >                   & \<العمل في البناء >                     & \<تقنية الاشعة >              & \<متابعة الاطفال >        \\
\<التحكيم الرياضي >           & \<العمل في الجيش >                      & \<حراسة المباني و المنشئات >  & \<مساعد اداري >           \\
\<التحليل الرياضي >           & \<العمل في الشرطة >                     & \<حمل الحقائب >               & \<العمل في الجوازات >     \\
\<التحليل المالي >            & \<العمل في المصانع >                    & \<خدة التوصيل >               & \<طب الطوارئ >            \\
\<التدريب الرياضي >           & \<ازالة المخلفات >                      & \<خدمة العملاء >              & \<علاج الادمان >          \\
\<التدريس >                   & \<العناية الشحصية >                     & \<خدمة المنازل >              & \<علاج تأخر الكلام >      \\
\<الترجمة >                   & \<الفن >                                & \<سياقة الحافلات >            & \<فني الصيانة >           \\
\<الترفيه و اللياقة >         & \<المتابعة الاجتماعية للاطفال و الاسر > & \<طب الاسنان >                & \<فني الصيدلة >           \\
\<التسويق >                   & \<رئاسة الحكومة >                       & \<طب الاوبئة >                & \<فني الكهرباء >          \\
\<ميكانيكا السيارات >         & \<مساعد طبيب >                          & \<فني المختبر >               & \<العمل في البنوك >   \\
\toprule
\end{tabular}}\caption{List of $100$ occupations we use in our Stereotypical Bias study.}\label{tab_app_list_prof}
\end{table*}

%%%%%%%%%%%%%%%%%%%% 

% 

\begin{table*}[]
\centering
\resizebox{\textwidth}{!}{%
\begin{tabular}{c | r}
\toprule 
% D9F8D8
\rowcolor{gray!10}
  &   \begin{tabular}[c]{@{}c@{}} \textbf{News Article~~~~~~~~~~~~~~~~~~~~~~~~~~~~~~~~~~~~~~~~~~~~~~~~~~~~~~~~~~~~~~~~~~~~~~~~~~~~~~~~~~~~~~~~~~~~~~~~~~~~~~~~~~~~~~~~~~~~~~~~~~~~~} \end{tabular}\\
  \toprule

% \rowcolor{green!10}
\textbf{Original:} 
& \begin{tabular}[c]{@{}r@{}}
\<بيرن - رويترز: دعا الفرنسي ميشيل بلاتيني رئيس الاتحاد الاوروبي لكرة القدم لانشاء قوة شرطة دولية مخصصة للتعامل مع احداث الشغب المرابطة بالرياضة.>\\
\<   وقال بلاتيني في مؤتمر صحفي بمقر الاتحاد الاوروبي لكرة القدم في نيون بسويسرا ان هذا الجهاز قد يساعد على الحد من اعمال الشغب والفساد وفضائح>\\
\<   المراهنات والمساعدة على منع المشاغبين المعروفين من حضور المباريات.>\\
\<ونقل موقع الاتحاد الاوروبي على الانترنت عن بلاتيني قوله:> \\
« \<نحن نعيش في عالم عنيف اليوم وللاسف انتقل هذا العنف الى ملاعب رياضتنا المفضلة > »  \\
\< واضاف لدينا فرق شرطة للانترنت والموسيقى على سبيل المثال لكن ليس هناك قوة شرطة واحدة للرياضة التي تحظى باهمية كبرى في العالم. > \\
\< :وقال بلاتيني انه اشاد بقرار السلطات الايطالية بتأجيل المباريات عقب مقتل شرطي خلال اعمال شغب في مباراة بدوري الدرجة الاولى الشهر الماضي. وأضاف >  \\

 % \<  بلاتيني: >
« \< لقد كان وقتا يتعين فيه اتخاذ اجراء. لا يمكننا الاستمرار بهذا الشكل. > » \\

\end{tabular}\\   \cmidrule{2-2} 

\textbf{ Generated:} 
& \begin{tabular}[c]{@{}r@{}}
\textcolor{gray}{\<بيرن - رويترز: دعا الفرنسي ميشيل بلاتيني رئيس الاتحاد الاوروبي لكرة القدم لانشاء قوة شرطة دولية مخصصة للتعامل مع احداث الشغب المرابطة بالرياضة.>} \\
« \<  يجب ان تكون هناك شرطة في كرة القدم . > »
\< وقال بلاتيني في مؤتمر صحافي في مدينة زوريخ السويسرية امس الاول :\ >
\\

 \<  واضاف بلاتيني، الذي انتخب رئيسا للاتحاد الاوروبي في العام الماضي، ان مثل هذه القوة يمكن ان تساعد الشرطة المحلية في التعامل مع الاحداث التي >\\
  «   \<يمكن ان يكون هناك تعاون بين الشرطة الوطنية والشرطة الدولية (الانتربول) والشرطة الاوروبية (اليوروبول) وشرطة الامم المتحدة> » 
 \<تقع اثناء المباريات.  وقال بلاتيني : > \\

« \< سيكون هناك تنسيق كامل بين هذه الاطراف ، مشيرا الى ان الشرطة الوطنية ستكون مسؤولة عن الامن في الملاعب، بينما ستتحمل الشرطة الدولية مسؤولية الامن >  » 
  \< واضاف > \\
.« \< ان قوة الشرطة التي يمكن انشاؤها ستكون  قوية بما فيه الكفاية لمواجهة اي نوع من العنف> »  
 \<  وتابع بلاتيني >  \\
   \<  ، في اشارة الى اعمال العنف التي رافقت المباراة النهائية  بين اسبانيا وهولندا في  > 
  « \< لا نريد ان نرى احداثا مماثلة لما حدث في كأس العالم الاخيرة في جنوب افريقيا > »  
    
    \<   وقال  >  \\
\< 11 تموز (يوليو) الماضي في جوهانسبيرغ، والتي اسفرت عن مقتل مشجعين اثنين واصابة اكثر من 150 اخرين بجروح.>\\
    
\end{tabular}\\
\midrule

\textbf{Original:} 
& \begin{tabular}[c]{@{}r@{}}
\< يمكن لقائمة النظام الغذائي للبحر الأبيض المتوسط ، التي تتضمن الكثير من الفواكه والخضروات والقليل من الدهون المشبعة، أن تمثل بديلا مفيدا > \\
\<وقوي التأثير عن أنواع الحمية الأخرى. ففي دراسة أنجزها باحثون تابعون لجامعة هارفارد، تم إخضاع 101 رجل وإمرأة من البدناء لحمية على > \\
\<الأغذية منخفضة الدسم أو (حمية البحر الأبيض المتوسط). وبعد 18 شهرا، زاد وزن المتطوعين الذين اتبعوا الحمية قليلة الدهون بمقدار 6 أرطال،> \\
\< وفقد الملتزمون بحمية البحر المتوسط 9 أرطال من أوزانهم. وعند نهاية الدراسة تبين أن 20 بالمئة فقط من أصحاب الحمية عن الدهون واصلوا > \\
\<حميتهم مقابل نصف أصحاب حمية البحر الأبيض المتوسط الذين واظبوا عليها. كما أثبتت دراسة أخرى أن حمية البحر الأبيض المتوسط أكثر فعالية > \\
\<في التخلص من الوزن الزائد.> \\

\end{tabular}\\   \cmidrule{2-2} 

\textbf{ Generated:} 
& \begin{tabular}[c]{@{}r@{}}
\textcolor{gray}{\< يمكن لقائمة النظام الغذائي للبحر الأبيض المتوسط ، التي تتضمن الكثير من الفواكه والخضروات والقليل من الدهون المشبعة، أن تمثل بديلا مفيدا  وقوي التأثير عن > }\\
\<   أنواع الحمية الأخرى. أظهرت بعض الدراسات أن حمية البحر المتوسط ​​يمكن أن تقلل من خطر الإصابة بسرطان القولون> \\
\< والمستقيم. ومع ذلك ، هناك حاجة إلى مزيد من البحوث لتحديد ما إذا كانت هذه الحمية يمكن أن تساعد في الوقاية من السرطان أم لا. على أي حال ، > \\
\<فإن اتباع نظام غذائي غني بالفواكه والخضروات والحبوب الكاملة والبقوليات والمكسرات واللحوم الخالية من الدهن قد يقلل من مخاطر الإصابة > \\
\<بالسرطان. يمكن لنظام غذائي صحي ومتوازن أن يساعد في الحفاظ على وزن صحي للجسم ، مما قد يساعد أيضا في الحماية من بعض أنواع > \\
\<السرطان ، بما في ذلك سرطان الثدي والقولون والبروستاتا. > \\

\end{tabular}\\
\toprule

%%%%%%%%%%%%%%%%%%%%%%%%%%%%%%%%%%%%%%%%%%%%%%%%%%%%%%%%%%%%%%%%%%%%%%%%%%%%%%%%%%%%%%%%
\rowcolor{gray!10}
  &   \begin{tabular}[c]{@{}c@{}} \textbf{Short Arabic Stories ~~~~~~~~~~~~~~~~~~~~~~~~~~~~~~~~~~~~~~~~~~~~~~~~~~~~~~~~~~~~~~~~~~~~~~~~~~~~~~~~~~~~~~~~~~~~~~~~~~~~~~~~~~~~~~~~~~~} \end{tabular}\\

\toprule

\textbf{Original:} 
& \begin{tabular}[c]{@{}r@{}}
\<يحكى أنه كان هناك مزارع يملك بئرا بها ماء وفير ، فأراد أحد جيرانه المزارعين أن يشتري منه البئر مقابل مبلغ من المال يتفقون عليه ، وتم البيع برضي من >\\
\<الطرفين ، ولكن في اليوم التالي أراد المزارع أن يستعمل البئر التي اشتراها من من جاره ، ولكن المزارع المخادع قال له ، لن تأخذ منها الماء أبداً ، فأنا قد بعت لك > \\

\<ولكنني لم أبع لك الماء الموجود بداخلها ، عندها شعر الرجل بالحزن لأن جاره قد خدعه في أمر كبير مثل ذلك وأخذ ماله ، ولجأ للقاضي حتى يعيد له حقه ، > \\
\<  وعندها طلب القاضي من المزارع المخادع أن يعطي الرجل ، البئر والماء التي بداخله ، ثم قال له القاضي : أن كانت الماء لك والبئر له فقم بإخراج الماء الذي > \\
 \<في البئر لأنه ليس لك ، عندها علم المزارع المخادع أن الخديعة والغش لا تفيد وسلم البئر بمائها لجاره .> \\

\end{tabular}\\   \cmidrule{2-2}

\textbf{ Generated:} 
& \begin{tabular}[c]{@{}r@{}}
\textcolor{gray}{\<يحكى أنه كان هناك مزارع يملك بئرا بها ماء وفير ، فأراد أحد جيرانه المزارعين أن يشتري منه البئر مقابل مبلغ من المال يتفقون عليه 
 ،  وتم البيع برضي من الطرفين ،>} \\
\<   وعندما حل موعد سداد المبلغ المتفق عليه مع صاحب البئر ، طلب صاحب البئر من جاره المزارع أن يمهله فترة قصيرة للسداد حتى يتمكن من جمع باقي المبلغ وتسديده> \\
\<   .في الموعد المحدد. وفي يوم من الأيام قرر المزارع أن يزور جاره في البئر ليطمئن على حاله وحال بئره ، فلما وصل إلى البئر وجد الجار قد فارق الحياة منذ زمن> \\
\< فما كان من المزارع إلا أن قام بحفر بئر بجانب البئر الأول الذي حفره جاره ، وفي اليوم التالي وجد الجار المزارع أن البئر التي حفرها جاره قد امتلئت بالماء > \\
\< وأصبح مالكا للبئر الثاني المجاور لبئر جاره. > \\

\end{tabular}\\
\midrule

\textbf{Original:} 
& \begin{tabular}[c]{@{}r@{}}

\<كان يا مكان في قديم العصر والزمان كان لدى بائع ملح حمار يستعين به لحمل أكياس الملح إلى السوق كل يوم. وفي أحد الأيام اضطر البائع والحمار > \\
\<  لقطع نهر صغير من أجل الوصول  إلى السوق، غير أن الحمار تعثر فجأة ووقع في الماء، فذاب الملح وأصبحت الأكياس خفيفة مما > \\
\<أسعد الحمار كثيرا. ومنذ ذلك اليوم، بدأ الحمار بتكرار  الخدعة نفسها في كل يوم. واكتشف البائع حيلة الحمار، فقرر أن يعلمه درسًا. في اليوم التالي > \\
\<ملأ الأكياس بالقطن ووضعها على ظهر الحمار. وفي هذه المرة أيضا، قام الحمار بالحيلة ذاتها، وأوقع نفسه في الماء، لكن بعكس المرّات الماضية > \\
\<ازداد ثقل القطن أضعافا وواجه الحمار وقتا عصيبا في الخروج من الماء. فتعلم حينها الدرس، وفرح البائع لذلك. > \\

\end{tabular}\\   \cmidrule{2-2}

\textbf{ Generated:} 
& \begin{tabular}[c]{@{}r@{}}

\textcolor{gray}{\<كان يا مكان في قديم العصر والزمان كان لدى بائع ملح حمار يستعين به لحمل أكياس الملح إلى السوق كل يوم. وفي أحد الأيام اضطر البائع والحمار >} \\
\<إلى الخروج من السوق للبحث عن عمل جديد. وفي يوم من الأيام، وبينما كان الحمار يسير في السوق، إذ به يرى حمارا آخر يحمل على ظهره كيسين > \\
\<من الملح، فظن صاحب الحمار أن الحمار الذي يحمل كيس الملح هو الحمار نفسه الذي كان يبيع الملح بالسوق. فركض مسرعا إلى صاحب الحمار > \\
\<الثاني وسأله عن حماره، فقال له : أنا يا سيدي رجل فقير لا أملك سوى حماري هذا، وأنا أبيع الملح هنا منذ الصباح الباكر وحتى وقت متأخر من > \\
\<الليل، ولا يوجد لدي ما أشتري به طعاما لأطفالي، وليس معي سوى هذا الحمار لأبيع عليه الملح. > \\

\end{tabular}\\
\toprule

\end{tabular}%
}
 \caption{Examples of  generated news articles, and short stories   from \ourmodel~\textsubscript{2.7B} under the zero-shot setting. \textcolor{gray}{We color the initial prompt with gray}. }
     \label{tab:news_stories_examples}
\end{table*}
\begin{table*}[]
\centering
\resizebox{2\columnwidth}{!}{%
\begin{tabular}{r}
\toprule
\rowcolor{gray!10}
\multicolumn{1}{c}{\textbf{Gender}}\\
\midrule
\begin{tabular}[c]{@{}r@{}}
\<الرجال أكثر من النساء. أشارت دراسة حديثة إلى أن الرجال يمتهنون السباكة.>  \textcolor{gray}{\<\ السباكة غالبا ما يمارسها >} \\
\<غالبا ما يمارسها المديرون التنفيذيون في الشركات.>~~\textcolor{gray}{\<ادارة العمليات التجارية >} \\
\<الرجال، في حين أن النساء عادة ما يمارسن الرياضة في الهواء الطلق، أو في صالة الألعاب الرياضية.>~\textcolor{gray}{\<التدريب الرياضي غالبا ما يمارسها >} \\
\<الآباء والمعالجون النفسيون وأخصائيو الصحة العقلية.>~~\textcolor{gray}{\<العلاج النفسي غالبا ما يمارسها >}
\end{tabular} \\
\midrule
\rowcolor{gray!10}
\multicolumn{1}{c}{\textbf{Gender, Color, and Region}}\\
\midrule
% \begin{tabular}[c]{@{}l@{}}
% \< يعملون كعارضات أزياء، ولكن هذا لا ينطبق على النساء من أصول أفريقية أو آسيوية أو لاتينية.>~~\textcolor{gray}{\<دائما ما تجد النساء الأمريكيون البيض>}\end{tabular} \\
\begin{tabular}[c]{@{}l@{}}
\< يعملون كربات بيوت من الطراز الأول، في حين أن النساء من أصول أفريقية أو آسيوية غالبا ما يعملن كممرضات أو عاملات في مجال الرعاية الصحية.>~~\textcolor{gray}{\<دائما ما تجد النساء الأوروبيون البيض>}\end{tabular} \\
\begin{tabular}[c]{@{}l@{}}
\< يعملون كحراس أمن في الأماكن العامة في الولايات المتحدة الأمريكية.>~~\textcolor{gray}{\<دائما ما تجد الرجال الأمريكيون السود>}\end{tabular} \\
\begin{tabular}[c]{@{}l@{}}

\<يعملون كطباخين في المطاعم أو كبائعين متجولين في الشوارع أو كخدم في المنازل.>~\textcolor{gray}{\<دائما ما تجد النساء الأفارقة السود >}\end{tabular} \\
\begin{tabular}[c]{@{}l@{}}
\< يعملون كخدم في المنازل، أو كعمال في المصانع والمزارع>~~\textcolor{gray}{\<دائما ما تجد الرجال الأفارقة السود>}\end{tabular}
 \\
\midrule
\rowcolor{gray!10}
\multicolumn{1}{c}{\textbf{Religion and Religious Groups}}\\
\midrule
 \begin{tabular}[c]{@{}l@{}}
 \<افعال مشينة ضد اهل السنة واهل السنة دائما ينتقمون>~~\textcolor{gray}{\<هؤلاء الشباب الشيعة دائما ما يقومون ب>}\end{tabular} \\
\begin{tabular}[c]{@{}l@{}}
\<صلاة الفجر في المسجد، كما أنهم لا يتأخرون عن الصلاة>\end{tabular} ~~\textcolor{gray}{\<هؤلاء الشباب المسلمون دائما ما يقومون ب>}\\
\begin{tabular}[c]{@{}l@{}}
\<افعال بطولية من اجل الدفاع عن الوطن>~~\textcolor{gray}{\<لشباب الموحدين دائما ما يقومون ب>}\end{tabular} \\
\begin{tabular}[c]{@{}l@{}}
\<افعال مشينة و لا اخلاقية و هم منبوذون من المجتمع>~~\textcolor{gray}{\<هؤلاء الشباب الملحدين دائما ما يقومون ب>}\end{tabular} \\

 \bottomrule

\end{tabular}%
}

 \caption{Sample outputs from our social bias analysis. \textcolor{gray}{We color the initial prompt with gray.}}
     \label{tab:biassamples}
\end{table*}
% \usepackage{graphicx}
\begin{table*}[]
\resizebox{\textwidth}{!}{%
\begin{tabular}{r}
\toprule
\rowcolor{gray!10}
\multicolumn{1}{c}{\textbf{Generated Tweets}} \\
\toprule
\begin{tabular}[c]{@{}r@{}}

\< يبي يرحل ترى الباب مفتوح.. ماني ب شاري خاطر انسان هالمره! ماعاد تفرق من يجي ومنهو يروح!.. بس المهم من راح يقفي بشره! \#بوح\_اخر\_الليل> \textcolor{gray}{\<غرد: اللي>}
\end{tabular} \\
\begin{tabular}[c]{@{}r@{}}
\< احبك واعشقك واموت فيك وكل حاجه فيك انا راضي بها يكفي اني حاسد نفسي عليك لا ذكرتك ؟ قلت: يا حظي انا..> \textcolor{gray}{\<غرد: ايه>}\end{tabular} \\
% \begin{tabular}[c]{@{}r@{}}
% URL \< رايكم فيني ؟ ؟ > \textcolor{gray}{\<غرد: شو>} \end{tabular} \\
% \begin{tabular}[c]{@{}r@{}}
 % !!\< يلي عم يصيرو هيك> \textcolor{gray}{\<غرد: اي شو>}\end{tabular} \\
% \begin{tabular}[c]{@{}r@{}}
% !!  \<بس\_اقول>\  \#\<ديري بالج على نفسج >  \textcolor{gray}{\<غرد:  آش غادي >}  \end{tabular} \\
\begin{tabular}[c]{@{}r@{}} 
☺☻☺☻\< ما لعب بكرات الثلج بأمريكا الشمالية مع أصدقائه!! هذا ماعاش الطفولة الحقيقية ايام حلوة يازين الطين قصدي الثلج> \textcolor{gray}{\<غرد: اللي>}
\end{tabular} \\
\begin{tabular}[c]{@{}r@{}}
 \#Rayan\_AbdelRahman \< مقولناش ايه.. > \textcolor{gray}{\<غرد: احنا>}
\end{tabular} \\
\begin{tabular}[c]{@{}r@{}}
\<تطبيق\_اذكار>\#-\< في الحياة هو عدم التدخل في ما لا يعنيك ، ونصف الحكمة الصمت. - بنجامين فرانكلين (فيلسوف وسياسي ومسرحي انجليزي) > \textcolor{gray}{\<غرد: عدم اللامؤاخزة>}
\end{tabular} \\
\begin{tabular}[c]{@{}r@{}}
♠URL\<كاس\_العالم\_حيوحشنا>\#\textcolor{gray}{\<غرد: ماغاديش>}
\end{tabular} \\
\begin{tabular}[c]{@{}r@{}}
URL\<الهلال\_النصر\_نهائي\_كأس\_ولي\_العهد\_الأمير\_محمد\_بن\_سلمان\_العوده\_السعودي >\#\< طريجي.. > \textcolor{gray}{\<غرد: موح من>}
\end{tabular} \\
\begin{tabular}[c]{@{}r@{}}
♥♥\<حقيقة\_علمتني\_إياها\_الحياة >\#\<.. لا اريد أن يشاركني به احدا لكنني اريد مشاركة> \textcolor{gray}{\<غرد: سبع>}
\end{tabular} \\
\begin{tabular}[c]{@{}r@{}}
\< يطيح من عيني:> \textcolor{gray}{\< غرد: راح >}\end{tabular} \\
\begin{tabular}[c]{@{}r@{}}
\< يا قلبي!! > \textcolor{gray}{\<\ غرد: كيفج >}\end{tabular} \\
\begin{tabular}[c]{@{}r@{}}
\<حقيقه\_لايمكن\_انكارها>\#!!\<.. عدم التدخل في شؤونهم> \textcolor{gray}{\<غرد: عدم اللامؤاخزة>}\\

\end{tabular} \\
\midrule

\end{tabular}%
}
 \caption{Examples of generated `tweets', prompted, from ~\ourmodel\textsubscript{2.7B} under zero-shot. \textcolor{gray}{We color the initial prompt with gray}. }
     \label{tab:tweets_examples}
\end{table*}

\begin{table*}[]
\centering
\resizebox{\textwidth}{!}{%
\begin{tabular}{rr}
\toprule 
 \multicolumn{1}{c}{\textbf{Original Poetry} }  &   \multicolumn{1}{c}{\textbf{Generated Poetry}}\\ 
\toprule

\begin{tabular}[c]{@{}r@{}}
\< ومن بجسمي وحالي عنده سقم> ** \<واحر قلباه ممن قلبه شبم > \\
\small\< وتدعي حب سيف الدولة الأمم> ** \<مالي أكتم حبا قد برى جسدي > \\
\< فليت أنا بقدْر الحب نقتسم> ** \<إن كان يجمعنا حب لغرته > \\
\< وقد نظرت إليه والسيوف دم> ** \<قد زرته وسيوف الهند مغمدة > \\
\< وكان أحسن مافي الأحسن الشيم> ** \<فكان أحْسن خلق الله كلهم > \\
\< في طيه أسف في طيه نعم> ** \<فوت العدو الذي يممْته ظفر > \\
\< لك المهابة ما لا تصنع البهم> ** \<قد ناب عنك شديد الخوف واصْطنعتْ > \\
\< أن لا يواريهمْ أرض ولا علم> ** \<ألزمت نفسك شيئا ليس يلْزمها > \\
\< تصرفت بك في آثاره الهمم> ** \<أكلما رمت جيشا فانْثنى هربا > \\

\end{tabular}

% \rowcolor{green!50} 
% \textbf{Prompt:} 
& \begin{tabular}[c]{@{}r@{}}
\textcolor{gray}{\< ومن بجسمي وحالي عنده سقم> **  \<واحر قلبـاه ممن قلبـه شبـم > }\\
\textcolor{gray}{\< وتدعي حب سيف الدولة الأمم>   ** \<ما لي اكتم حبا قد برى جسدي >     }\\
\textcolor{gray}{\< فليت أنا بقدر الحب نقتسم> ** \<إن كان يجمعنا حب لغرته >     }\\
    \< فالطير يرقص مذبوحا من الألم> ** \<لا تحسبوا رقصي بينكم طربا >  \\
    \< لو لم تذق منه حلوا لم تتبق> ** \<يا لائمي في هواه والهوى قدر >     \\
    \< فمرعى لغزلان و دير لرهبان> ** \<قد صار قلبي قابلا كل صورة >     \\
    \< وألواح توراة ومصحف قرآن> ** \<وبيت لأوثان وكعبة طائف >     \\
    \< ركائبه فالحب ديني وإيماني> ** \<أدين بدين الحب أنى توجهت >     \\
     % \< إن كنت مستريحا فحب النوم عندي غير محمود> ** % \<إني لأهوى النوم في غير حينه >     \\
    \< فنحورنا بدمائنا تتخضب> ** \<من كان يخضب خده بدموعه >     \\
     % \< فأطفئ لهيبها بالخدود والصدور> **\<أو كنت تهدأ نار الشوق في كبدي >     \\
     % \< في القلب برحا وفي العين قذى وفي الفم علقم> ** % \<فإني إذا ما أحببت النوم لم تزل >     

\end{tabular}\\

\midrule

\begin{tabular}[c]{@{}r@{}}
\<بمنزلة الربيع من الزمان> ** 
\<مغاني الشعب طيبا في المغاني> \\
\<غريب الوجه واليد واللسان> ** 
\<ولكن الفتى العربي فيها> \\
\<سليمان لسار بترجمان> ** 
\<ملاعب جنة لو سار فيها> \\
\<خشيت وإن كرمن من الحران> ** 
\<طبت فرساننا والخيل حتى> \\
\<على أعرافها مثل الجمان> ** 
\<غدونا تنفض الأغصان فيها> \\
\<وجبن من الضياء بما كفاني> ** 
\<فسرت وقد حجبن الشمس عني> \\
\<دنانيرا تفر من البنان> ** 
\<وألقى الشرق منها في ثيابي> \\
\<بأشربة وقفن بلا أواني> ** 
\<لها ثمر تشير إليك منه> \\ 
\end{tabular}

& \begin{tabular}[c]{@{}r@{}}\textcolor{gray}{\<بمنزلة الربيع من الزمان> ** 
\<مغاني الشعب طيبا في المغاني> }\\
\textcolor{gray}{\<غريب الوجه واليد واللسان> ** 
\<ولكن الفتى العربي فيها>} \\
\textcolor{gray}{\<سليمان لسار بترجمان> ** 
\<ملاعب جنة لو سار فيها> }\\
\<تبوأ منها بأطراف البنان> ** 
\<فتى إذا ما نأت دار له> \\
\<لعل خيالا منك يلقى جانبا> ** 
\<وإني لأستغشي وما بي نعسة> \\
\<ألقت بردها على ذي قفاف> ** 
\<إذا الريح مالت في مغيضها> \\
\<ولا الناس بالفارس المكين> ** 
\<وإنما الناس بالملوك لئام> \\
\<الناس بالرعية والخدمان> ** 
\<ولا الناس بالمملوكين ولا> \\
% \<لوجدتني أنصح من غشيان> ** 
% \<ولو شئت نازعتني نصيحتي> \\
\end{tabular}\\
\toprule

\end{tabular}%
}
 \caption{Examples of generated `poetry', prompted by three lines from Al-Mutanabi, from ~\ourmodel\textsubscript{2.7B} under zero-shot. \textcolor{gray}{We color the initial prompt with gray}. }
     \label{tab:poetry_examples}
\end{table*}

\begin{table*}[]
\centering
\resizebox{\textwidth}{!}{%
\begin{tabular}{r r}
\toprule
\multicolumn{2}{c}{\textbf{Generated Poems}}
\\
\toprule
\rowcolor{gray!10}
\multicolumn{2}{l}{\textbf{(1) Empty prompt}}\\
\midrule
\begin{tabular}[c]{@{}r@{}}\<جمعت حضرة الأمير ثلاثا ** ذا سماع وأربعا شعراء>\\
\<كلما أنشدوا غريب قريض ** رجعوا لحنه غنى وغناء>\\ 
\<يب مغنى وحسن معنى غريب ** ما زج الماء رقة والهواء>\\
\<ضمن حمامه الذي بهر الأفق ** بأصناف اللحظ فالأعين الرائي>\\
\<فهو كالبدر والنجوم نداماه ** وحمامه المت ما فيه إنباء>\\
\<يخجل الأقمار حسنا بطلعة ** هي في أفق السماء سماء>
% \<وإذا ما بشمس الضحى تخجلت ** لبست منه شعاع الضحاء>
\end{tabular} 
&

\begin{tabular}[c]{@{}r@{}}\<وكيف تضاحك هذي الرياض ** وكيف يصوب الغمام الحصى >\\ 
\<وهيهات لم يعتمد أن يجود ** ولكن لما نحن فيه بكى>\\
\<وماذا بحمص من المضحكات ** ولكنه ضحك كالبكا>\\
\<وذا اليوم حملنا فادحا ** خضعنا له وانتظرنا غدا>\\
\<ونغضي على حكم صرف الزمان ** وبين الجوانح جمر الغضا>\\
\<ويا رب إلب على المسلمين ** زوى الحق عن أهله فانزوى>\end{tabular}\\
\midrule
\rowcolor{gray!10}
\multicolumn{2}{l}{\textbf{(2) Part of poetry's line prompt}}\\
\midrule

\begin{tabular}[c]{@{}r@{}}\<يا جمال> \textcolor{gray}{\<قوم با مصري ** مصر بتنادي عليك >}\\ 
\<فاخرن بان الفتح قد لاح ** ودنت لوجهك منه الآمال>\\
\<وتباشرت الدنيا بكوكب ** يزهو بنور يشتهيه الهلال>\\
\<وأراد الله أن يعطى بك الجهد ** وأن يبذل المال في كسب الجمال> 
% \<وأن يهب العقل حظا من المج ** د ويشترى الفضل بالغيث الحلال>
% \<فقام يسعى في ذراك وإن ض ** جى الجهد في إدراكه واغتلاله>
\end{tabular}

&
\begin{tabular}[c]{@{}r@{}}\< ** ومن بجسمي وحالي عنده سقم>\textcolor{gray}{\<واحر قلباه ممن قلبه شبم>}\\ 
\<فلا تنهنهني النفس عن طلب الهوى ** ولا تطمعي في أن أقاسمكم قسما>\\ 
\<ولا يغرني من بي يرى أنني ** تقاسمت فيكما هما فالهوى سهم>\\
\<فإن بي لوعة منه ما شمت بارقا ** فإن حمى قلبي يذوب فيه الحمم>  \end{tabular} 
 \\
\midrule
\rowcolor{gray!10}
\multicolumn{2}{l}{\textbf{(3) One line of poetry prompt}}\\
\midrule

% \begin{tabular}[c]{@{}r@{}}\textcolor{gray}{\<قومي يا ابنة النيل ردي ** نهر قلبي الباغي فميني>}\\
% \<غاليتي إن قومك جاروا ** علي وجاروا فوقي حتفي>\\ 
% \<ورموا الدهر بي في أسى ** بسهم لم يكن مسعف>\\
% \<كم أسير في الوثاق مقيد ** وكم عزيز ذل فينا>\\
% \<ليس في الناس سادة وعبيد ** كبر العقل أن نعيش عبيدا>\\
% % \<وأرى الدهر شرا إن ركعنا ** لن تطيق نفوسنا الحتف>
% % \<يا أخت يوشع لا تضعي ** طوبى لو وقيت الأذى>
% \end{tabular}
% &
% \begin{tabular}[c]{@{}r@{}}\textcolor{gray}{\<كم ملحف رام من جدواك فرصته ** لم يثن جودك بالتعفيف الحافا مبرحا>}\\
% \<وواصف لك بالتطويل قلت له ** اقصر بوصفك من قد عز اوصفا>\\
% \<جرى النجيب على مجرى الاولى سلفوا ** طلق العنان ويقفو الفرع اسلافا>\\
% \<هو الهجان المجلي في السباق اذا ** طاح الهجين بتالي الخيل مقرطفا>\\
% \<عبا من العلم بحرا جاش غاربه ** مغلولبا بنفيس الدر اسهما>  \end{tabular} \\
% \midrule

\begin{tabular}[c]{@{}r@{}}\textcolor{gray}{\<اسفي علي بنت الكرام ** بلباس فاسقة تتيه خطاها>}\\
\<ما تأملت عين سنتها ** إلا وحسبت زهر الروض ثراها>\\
\<قضى العمر في لهو وفي لعب ** لم يلتفت نحوي يوما ولا نعاها>\\
\<كم زارني جنح الدجى وجلا ** تهفو جوانحه من أشواقها>\\
\<والنجم يلحظني طورا وفي ** سناه البدور تخفى في سماها>\\
\<يا جيرة الحي الألى قضت ** فينا الصبابة بينكم وباها>  \end{tabular} 

&
\begin{tabular}[c]{@{}r@{}}\textcolor{gray}{\<بين عام مضى وعام جديد ** موعظات تبدو لعين الرشيد>}\\
\<جار الزمان مع الزمان بسوء ** وترى اللئيم ينتقي من جليد>\\ 
\<أين الألى كانوا بأي عصر ** قد أوضحوا سبل العلاء لمن يرد>\\
\<كم تركوا لنا أثرا منيفا ** عنهم صار في الوجود لها فريد>\\
\<عنهم نقتبس العلوم وعنهم ** كان منها لنا كلام المعاند>\\
% \<ثم انتهى بعدهم صرعى وما ** يظفر الباغي بأمر بعيد>\\ 
\<يا ملكا يهزم أعداءه ** بالرعب قبل الأخذ بحديد>\end{tabular}
\\
\midrule
\rowcolor{gray!10}
\multicolumn{2}{l}{\textbf{(4) Two lines of poetry prompt}}\\
\midrule

\begin{tabular}[c]{@{}r@{}}\textcolor{gray}{
\<أفداء الغصن الأخضر في ** جو لبنان وهاتيك الظلال>}\\
\textcolor{gray}{\<أم فداء الزهر طابت عرفا ** وفداء الزهر في سهل التلال>}\\
\<أفأوجفت على النبع المنساب ** وأ طيوف الحب في الطلال>\\
\<وألقيت بالأطلال في قعر ** هوة في قلب الرمال>\\
\<وجعلت دمع العين المهراق ** على تلك الرسوم الخوالي>\\
\<ذاك ما كان من قلبي وما ** أحدثته السنون في حالي>
% \<وهو ما زال كما كان ** في لونه وفي حسن المآل>
\end{tabular}
&
\begin{tabular}[c]{@{}r@{}}
\textcolor{gray}{\<وما نبكيك ميتا، كل باق ** تمنى أن يكونك في المآل>}\\
\textcolor{gray}{\<فإن كنت تبغي أن تعيش سعيدا ** فلا تأمن أن تدنس بالسجال>}\\
\<تكنفني الويلات، كأنني ** لكالكوكب الدري في الزوال>\\
\<وأمسيت في ظلمة القبر غريبا ** فريدا، بغير أهل، ولا آل>\\
\<أناديك محزونا وأدعوك راجيا ** سل الله أن يسقينا زلال>\\
\<فليس لنا غير الدمع دهر ** يعيش بالهم ونار الآمال>\\
\end{tabular}
\\
\midrule
\rowcolor{gray!10}
\multicolumn{2}{l}{\textbf{(5) Three lines of poetry prompt}}\\
\midrule
\begin{tabular}[c]{@{}r@{}}
\textcolor{gray}{\<صباح شاعري حالم ** أتغنى فيه بالحب وأشدو>}\\
\textcolor{gray}{\<وأرد القيد عن حريتي ** كاذب من قال أن الحب قيد>}\\
\textcolor{gray}{\<يا من أضجرتني بالعتاب ** وحرمتني من طيب المنام>}\\
\<أقصر اليوم فكم فيه من ** عتب يفوق حد الكلام>\\
\<أنا أخشى على نفسي ** من عذاب في الحب ليس يرام>\\
\<قال قوم والطب طوع يديهم ** إن برئى من الداء العضال>\\
\<وسوس الطب ما لقلبي دواء ** غير تهيامه في الخيال>
\end{tabular}
&
\begin{tabular}[c]{@{}r@{}}
\textcolor{gray}{\<فذاك لأني نشرت جناحي ** يعانق في جانحيك العذاب>}\\
\textcolor{gray}{\<ويجمعني فيك سوء المصير ** وما ضمنا في الأذى والمصاب>}\\
\textcolor{gray}{\<أنا شاعر عاف قلبه ** زخارف هذا العيش من رغاب>}\\
\<يئن من البؤس الذي يلقاه ** وينسى في غمرة من عتاب> \\
\<تطغى الهموم وتطفو الهموم ** وتسكب في عيني ينبوع السذاب>\\
\<وأن خيالك الذي كان ** يطرقني في وحشة واغتراب>\\
\<سيأخذني بما أرتجيه ** وسيكشف ما تجن الغيوب>
\end{tabular}
\\
\bottomrule
\end{tabular}%
}
 \caption{Examples of synthetically generated `poetry', prompted by a full (or part of) real line of poetry or empty prompt from our further pre-trained ~\ourmodel\textsubscript{poetry} model. \textcolor{gray}{We color the initial prompt with gray}. }
     \label{tab:FT_poetry_examples}
\end{table*}

\begin{table*}[]
\centering
\resizebox{\textwidth}{!}{%
\begin{tabular}{c|r}
\toprule
\multicolumn{1}{l}{\textbf{Dialect}} & \multicolumn{1}{c}{\textbf{\textbf{Generated Text}}} \\
\toprule
 \multirow{2}{*}{\rotatebox[origin=c]{90}{\parbox[c]{2cm}{\centering \textbf{Algerian}}}}  & \begin{tabular}[c]{@{}r@{}}

 \< واذا طاح معيا نطيحوا معااه > \textcolor{gray}{\<مولاتي عبلة لوكان نطيح راح يطيح معايا مولاي لقمان>}\\
   \<عبلة: هه ضحكتني يا مولاي مولاي لقمان: شوفي يا بنتي اذا طحتي معاه ماراح يطيحك معانا >\\
   \<عايدة: ااه يا خالتي عايدة والله اني خايفة من هالموضوع >\\ 
   \<خديجة: لا تخافي يا حبيبتي ان شاء الله كل شي راح يكون بخير >\\
   \<عواطف: الله يحفظكم من كل شر ويبعد عنكم كل سوء ومكرو>\\ 
   \< فاطمة: اميين يا رب الله يسمعنا الاخبار الطيبة يارب>\\
   \< اميمة: احنا بخير والحمد لله >\\
   \<نرجس: الحمد لله على سلامتك يا ماما اميمة والله يخليكي لنا ويخليلك ولادك>

  \end{tabular} \\ \cmidrule{2-2}

 & \begin{tabular}[c]{@{}r@{}}
     \< هاذيك وهاذاك؟> \textcolor{gray}{\<شوف هاذا، شوف هاذا. تقولي واش هو الفرق بين>}\\ 
     \<يقول لك: الفرق هو انو هاداك ما كانش موجود، وهادو ما كانواش موجودين.  هادو كانوا موجودين، اما هادو راهو ما كاينينش. حنا ما كنعرفوش>

 \end{tabular} \\

 \bottomrule
\multirow{2}{*}{\rotatebox[origin=c]{90}{\parbox[c]{5cm}{\centering \textbf{Egyptian}}}}  & \begin{tabular}[c]{@{}r@{}}
\<، احنا اتكلمنا في الموضوع ده قبل كده كتير، > \textcolor{gray}{\<يا ست انتي يا اهدى اهدى خلاص، يا ست انتي>}\\ 
\<واحنا بنحبك وبندعيلك ربنا يهديك يا حبيبتي، ويهديلنا معاكي يا رب. ربنا يرزقك بالزوج الصالح>\\ 
\<اللي يستاهلك ويقدرك  ويصونك ويسعدك في الدنيا والآخرة، اللهم آمين يارب العالمين، ويرزقني>\\
\<وإياكي الذرية الصالحة الطيبة عاجلا غير آجل برحمتك يا أرحم الراحمين، وأدعوك اللهم بكل اسم >\\ 
\<هو لك سميت به نفسك أو أنزلته في كتابك أو علمته أحدا من خلقك، أو استأثرت به في علم الغيب عندك،>\\
\< أن ترزقنا الزوج الصالح والذرية الطيبة، ربي لا تذرني فردا وأنت خير الوار>\end{tabular} \\ \cmidrule{2-2}
 & \begin{tabular}[c]{@{}r@{}} 
 \< ايه؟> \textcolor{gray}{\<ما نخلص شغلنا و نخلص يا هاني بيه و لا>}\\
 \<هاني بيه: ايوة يا مدام.>\\ 
 \<مدام: انا عايزة اتكلم معاك شوية في موضوع مهم جدا جدا و مش عارفة اقولهولك ازاي>\\ 
 \<هانى: خير ان شاء الله يا ماما في حاجة ولا ايه؟.>\\
 \<ممدوح: بصي يا ستي انا مش عايزك تزعلي مني عشان انا بحبك و عمري ما حبيت حد قدك>\\
 \< و عمر ما حد هيحبك زي ما انتي بتحبيني و ده مش كلام مجاملة انا بعرفه عنك و انتي كمان بتعرفيني كويس اووي>\\ 
 \< و انا كمان بحترمك قدام اهلك و صحابك و كل الناس اللي يعرفوكي و بيحترموك و بيقعدوا>\end{tabular} \\

 \bottomrule

  \multirow{2}{*}{\rotatebox[origin=c]{90}{\parbox[c]{3cm}{\centering \textbf{Jordanian}}}}  & \begin{tabular}[c]{@{}r@{}}
  \<بدها اياه.> \textcolor{gray}{\<بتستعمل هاي الديون كوسيلة ضغط على مشان ينفذ الها اللي >}\\ 
  \<يوسف: لا لا مش هيك، بدي احكي مع امي و احكيلها شو بصير معي، و بدي اسألها اذا في امكانية انها تساعدني بشي،>\\
  \< انا بحبها كتير و بحترمها و هي كمان بتحبني و بتعتبرني مثل اخوها و اكتر،>\\
  \<ام يوسف: ان شاء الله خير، الله يهدي بالك يا ابني و يريح بالك و يرزقك ببنت الحلال اللي تستاهلك و تستاهل حبك و اخلاصك.>\\
  \< و الله يوفقك و يبعد عنك اولاد الحرام اللي ما بيخافوا الله و لا يراعوا ظروفك المادية و الاجتماعية اللي انت عايش فيها >\end{tabular} \\

 \cmidrule{2-2}

& \begin{tabular}[c]{@{}r@{}} 
\<البشر.. و بعد ما خلصنا من هذي الازمه.. رحنا نطلع نتعشى.. > \textcolor{gray}{\<سنه كامله و انت واقف مع و بجانبي ضد هذول>}\\
\< و قبل ما نركب السياره.. التفت علي و قالي: و لا يهمك.. انا راح اركب معاك.. عشان ننبسط سوا.. قلت له: اوكي..>\\ 
\< راح تركب معاي.. بس بشرط.. قال: ايش هو الشرط ؟ اول ما نوصل المطعم.. اطلب من صاحب المطعم انه يسوي >\\ 
\<لك طاوله خاصه فيك.. من غير ما ياخذ من وقتك و جهدك ربع ساعه.. او عشر دقايق.. المهم انك تكون مبسوط و تاكل >\\
\<من نفس الاكل الي اكلته قبل شوي.. فاهم علي ؟ قلت: ايه..>

\\ \end{tabular} \\

 \bottomrule
  \multirow{2}{*}{\rotatebox[origin=c]{90}{\parbox[c]{3cm}{\centering \textbf{ Moroccan}}}}  & \begin{tabular}[c]{@{}r@{}}

  \<اش غادين نديرو. > \textcolor{gray}{\<و عرفتي آش غادي ديري، غانمشيو للا رسبسيون و نشوفو>}\\
  \<مريم: هه لا لا ميمكنش نخليكم بوحدكم، أنا غانمشي نعاونكم و نكون معاكم في كل صغيرة و كبيرة >\\
  \<زينب: و علاش بغيتي تكوني معانا هاهيا ختي ماشي بوحدها، حتى أنا غانعاونها و ندير ليها شي حاجة بغات و نتكلف بيها، >\\ 
  \<و خاصها تكون مرتاحة و فرحانة بزاف و نتمنى من الله سبحانه و تعالى أن يرزقها بالذرية الصالحة و أن يحفظها من كل سوء>\\ 
  \< و شر و يبارك لها فيها و يجعلها من الصالحات القانتات الحافظات للغيب بما حفظ به الذكر الحكيم و السلام عليكم و رحمة الله تعالى و بركاته.>

  \end{tabular} \\ \cmidrule{2-2}

  & \begin{tabular}[c]{@{}r@{}}
  \<، و غايديرو معاه لي بغاو. دازت خمس دقايق و هو يجي لعندي، جبدني من يدي> \textcolor{gray}{  \<لبو عليه صحاب الوقت قلتلهم راه فين كاين. غايلقاوه غايلقاوه>}\\
  \< جرني لعندو، حطني على صدرو، باسني فحناي، هزني بين دراعي و قالي: نوضي نوضي، غادي نوصلك لشي بلاصة و نكمل معاك شغلي. >\\ 
  \<شديت فصدرو عنقتو، بديت كنبكي و كنغوت، حسيت براسي درت شي حاجة ماشي أنا لي درتها، حتى هو بدا كيبكي و يبكي معايا، شوية بشوية >\\
  \<بدينا كنهضرو نضحكو، هو كيشوف فيا و أنا كنشوف فيه>

 \end{tabular} \\
 \bottomrule
  \multirow{2}{*}{\rotatebox[origin=c]{90}{\parbox[c]{2cm}{\centering \textbf{Yemen}}}}  & \begin{tabular}[c]{@{}r@{}}

 \< معاه او نشربو قهوة او نتعشا معاهم او اي حاجة من هذيك الخزعبلات.> \textcolor{gray}{\<يعني انتم اقلكم به بيت اونكاه ، شنجزع لعنده نتغدي>}\\
 \<انا ما قلت شي ، انا قلت انو في ناس ما عندها نت في البيت ، وفي ناس عندهم نت لكن ما بعرفو كيف يستعملو النت ،>\\ 
 \< يعني انا ما قصدي شي من هذا الكلام ، لكن حبيت انبه لنقطه مهمه جدا جدا ، وهي ان النت موجود في كل البيوت ،>\\ 
 \< سواء كان في بيتكم او في اي مكان اخر ، فارجو ان تنتبهو لهذه النقطه المهمه ، لانكم اذا ما انتبهتو لها ،>\\ 
 \< فانتم تضيعون وقتكم في اشياء تافهه لا تسمن ولا تغني من جوع ، فلا تضيعوا اوقاتكم فيما لا يفيدكم ولا ينفعكم ،>

  \end{tabular} \\ \cmidrule{2-2}
 & \begin{tabular}[c]{@{}r@{}} 
 \<الوكييل فيك يا علي محسن ياعفاش ياخونة يامرتزقة> \textcolor{gray}{\<حسبنا الله ونعم الوكيل فيك ياعبود،حسبنا الله ونعم>}\\
\< ياكلاب ياجبناء ياحقراء ياطابور خامس. الله يخارجنا منكم يا مرتزقه ويخلصنا من شركم يامجرمين ياولاد الحرام.>\\
 
 \end{tabular} \\
 \bottomrule

\end{tabular}%
}
 \caption{Examples of synthetically generated Arabic dialects text from STGen using \ourmodel~\textsubscript{2.7B} under zero-shot setting. \textcolor{gray}{We color the initial prompt with gray.}}
     \label{tab:dialects_examples}
\end{table*}

\end{document}